\documentclass[11pt,onecolumn,journal,draftcls]{IEEEtran}
\pdfoutput=1 

\usepackage{cite}

\usepackage[utf8]{inputenc} 
\usepackage[T1]{fontenc}    
\usepackage{url}            
\usepackage{booktabs}       
\usepackage{amsfonts}       
\usepackage{nicefrac}       
\usepackage{microtype}      

\usepackage{jurestyle}
\usepackage{juremath}
\usepackage{amssymb,amsmath}
\usepackage{bbm}
\usepackage{color}
\usepackage{xcolor}
\usepackage{graphicx}
\usepackage{subfigure}

\usepackage{pgfplots}
\pgfplotsset{compat=newest}

\usepackage{tikz}
\usepackage{textcomp}
\usetikzlibrary{plotmarks}
\usetikzlibrary{shapes,arrows}
\usetikzlibrary{backgrounds,calc,positioning}

\usepackage{amsthm}
\usepackage{centernot}

\newcommand{\review}[1]{{\color{black}#1}}

\title{Robust Large Margin Deep Neural Networks}

\author{Jure~Sokoli{\'c},~\IEEEmembership{Student Member,~IEEE,}
		Raja~Giryes,~\IEEEmembership{Member,~IEEE,}
		Guillermo~Sapiro,~\IEEEmembership{Fellow,~IEEE,}
		and~Miguel~R.~D.~Rodrigues,~\IEEEmembership{Senior Member,~IEEE}
		\thanks{J. Sokoli{\'c} and M. R. D. Rodrigues are with the Department of Electronic and Electrical Engineering, Univeristy College London, London, UK (e-mail:\texttt{ \{jure.sokolic.13, m.rodrigues\}@ucl.ac.uk}).}
		\thanks{R. Giryes is with the School of Electrical Engineering, Faculty of Engineering, Tel-Aviv University, Tel Aviv, Israel (e-mail: \texttt{raja@tauex.tau.ac.il}).}		
		\thanks{G. Sapiro is with the Department of Electrical and Computer Engineering, Duke University, NC, USA (e-mail: \texttt{guillermo.sapiro@duke.edu}).}		
\thanks{The work of Jure Sokoli\'c and Miguel R. D. Rodrigues was supported in part by EPSRC under grant EP/K033166/1. The work of Raja Giryes was supported in part by GIF, the German-Israeli Foundation for Scientific Research and Development. The work of Guillermo Sapiro was supported in part by NSF, ONR, ARO, and NGA. }
}

\newtheorem{theorem}{Theorem}

\newtheorem{corollary}{Corollary}
\newtheorem{lemma}{Lemma}
\newtheorem{definition}{Definition}
\newtheorem{remark}{Remark}

\begin{document}

\maketitle

\begin{abstract}
The generalization error of deep neural networks via their classification margin is studied in this work. Our approach is based on the Jacobian matrix of a deep neural network and can be applied to networks with arbitrary non-linearities and pooling layers, and to networks with different architectures such as feed forward networks and residual networks. Our analysis leads to the conclusion that a bounded spectral norm of the network's Jacobian matrix in the neighbourhood of the training samples is  crucial for a deep neural network of arbitrary depth and width to generalize well.  This is a significant improvement over the current bounds in the literature, which imply that the generalization error grows with either the width or the depth of the network. Moreover, it shows that the recently proposed batch normalization and weight normalization re-parametrizations enjoy good generalization properties, and leads to a novel network regularizer based on the network's Jacobian matrix. The analysis is supported with experimental results on the MNIST, CIFAR-10, LaRED and ImageNet datasets.
\end{abstract}


%

\section{Introduction} 

In recent years, deep neural networks (DNNs) achieved state-of-the-art results in image recognition, speech recognition and many other applications \cite{Krizhevsky2012,Hinton2012,LeCun2015,He2015}. DNNs are constructed as a series of non-linear signal transformations that are applied sequentially, where the parameters of each layer are estimated from the data \cite{LeCun2015}. Typically, each layer applies on its input a linear (or affine) transformation followed by a point-wise non-linearity such as the sigmoid function, the hyperbolic tangent function or the Rectified Linear Unit (ReLU) \cite{Nair2010a}. Many DNNs also include pooling layers, which act as down-sampling operators and may also provide invariance to various input transformations such as translation \cite{Bruna2013b,Boureau2010}. They may be linear, as in average pooling, or non-linear, as in max-pooling.

There were various attempts to provide a theoretical foundation for the representation power, optimization and generalization of DNNs. 
For example, the works in \cite{Cybenko1989,Hornik1991} showed that neural networks with a single hidden layer -- shallow networks --  can approximate any measurable Borel function. On the other hand, it was shown in \cite{Montufar2014a} that a deep network can divide the  space into an exponential number of sets, which can not be achieved by shallow networks that use the same number of parameters. Similarly, the authors in \cite{Cohen2015} conclude that functions implemented by DNNs are exponentially more expressive than functions implemented by shallow networks. The work in \cite{Telgarsky2016a} shows that for a given number of parameters and a given depth, there always exists a DNN that can be approximated by a shallower network only if the number of parameters in the shallow network is  exponential in the number of layers of the deep network.

Scattering transform - a convolutional DNN like transform, which is based on the wavelet transform and pointwise non-linearities - provides insights into translation invariance and stability to deformations of convolutional DNNs \cite{Mallat2012,Bruna2012,Wiatowski2015c}.

DNNs with random weights are studied in \cite{Giryes2015}, where it is shown that such networks perform distance preserving embedding of low-dimensional data manifolds. The authors in  \cite{Choromanska2015b} model a loss function of DNN with a spin-glass model and show that for large networks the local optima of the loss function are close to the global optima. Optimization aspects of DNNs are studied from the perspective of tensor factorization in \cite{Haeffele2015} where it is shown that if a network is large, then it is possible to find the global minima from any initialization with a gradient descent algorithm. The role of DNNs in improving convergence speed of various iterative algorithms is studied in~\cite{Giryes2016}.

Optimization dynamics of a deep linear network is studied in \cite{Saxe2013}, where it is shown that the learning speed of deep networks may be independent of their depth. Reparametrization of DNN for more efficient learning is studied in depth in \cite{Ollivier2015}. A modified version of stochastic gradient descent for optimization of DNNs that are invariant to weight rescaling in different layers is proposed in \cite{Neyshabur2015b}, where it is shown that such an optimization may lead to a smaller generalization error (GE) - the difference between the empirical error and the expected error, than the one achieved with the classical stochastic gradient descent. The authors in \cite{Ioffe2015} propose the batch normalization -- a technique that normalizes the output of each layer and leads to faster training and also a smaller GE. A similar technique based on normalization of the weight matrix rows is proposed in \cite{Salimans2016}. It is shown empirically  that such reparametrization leads to a faster training and a smaller GE. Learning of DNN by bounding the spectral norm of the weight matrices is proposed in \cite{An2015a}. Other methods for DNN regularization include weight decay, dropout \cite{Srivastava2014}, constraining the Jacobian matrix of encoder for regularization of auto-encoders \cite{Rifai2011}, and enforcing a DNN to be a partial isometry \cite{Huang2015}. 


An important theoretical aspect of DNNs is the effect of their architecture, e.g. depth and width, on their GE. Various measures such as the VC-dimension \cite{Vapnik1999,Shalev-Shwartz2014}, the Rademacher or Gaussian complexities \cite{Bartlett2003} and algorithmic robustness \cite{Xu2012a} have been used to bound the GE in the context of DNNs. For example, the VC-dimension of DNN with the hard-threshold non-linearity is equal to the number of parameters in the network, which implies that the sample complexity is linear in the number of parameters of the network. The GE can also be bounded independently of the number of parameters, provided that the norms of the weight matrices (the network's linear components) are constrained appropriately. Such constraints are usually enforced by training networks with weight decay regularization, which is simply the $\ell_1$- or $\ell_2$-norm of all the weights in the network. For example, the work \cite{Neyshabur2015} studies the GE of DNN with ReLUs and constraints on the norms of the weight matrices. However, it provides GE bounds that scale exponentially with the network depth. Similar behaviour is also depicted in \cite{Sun2015}. The authors in \cite{Xu2012a} show that DNNs are robust provided that the $\ell_1$-norm of the weights in each layer is bounded. The bounds are exponential in the $\ell_1$-norm of the weights if the norm is greater than 1.  

The GE bounds in \cite{Shalev-Shwartz2014,Neyshabur2015,Xu2012a} suggest that the GE of a DNN is bounded only if the number of training samples grows with the DNN depth or size. However, in practice increasing network's depth or size often leads to a lower GE \cite{He2015,Zagoruyko2016}.  Moreover, recent work in \cite{Zhang2016} shows that a 2 layer DNN with ReLUs may fit any function of $n$ samples in $d$ dimensions provided that it has $2n + d$ parameters, which is often the case in practice. They show that the nature of the GE depends more on the nature of the data than on the architecture of the network as the same network is able to fit both structured data and random data, where for the first the GE is very low and for the latter it is very large. The authors conclude that data agnostic measures such as the Rademacher complexity or VC-dimension are not adequate to explain the good generalization properties of modern DNN.

Our work complements the previous works on the GE of DNNs by bounding the GE in terms of the DNN classification margin, which is independent of the DNN depth and size, but takes into account the structure of the data (considering its covering number) and therefore avoids the issues presented above.  The extension of our results to invariant DNN is provided in  \cite{Sokolic2017b}.

\subsection{Contributions} 

In this work we focus on the GE of a multi-class DNN classifier with general non-linearities. We establish new GE bounds of DNN classifiers via their classification margin, i.e. the distance between the training sample and the non-linear decision boundary induced by the DNN classifier in the sample space. The work capitalizes on the algorithmic robustness framework in \cite{Xu2012a} to cast insight onto the generalization properties of DNNs. In particular, the use of this framework to understand the operation of DNNs involves various innovations, which include:

\begin{itemize}
 
	\item We derive bounds for the GE of DNNs by lower bounding their classification margin. The lower bound of the classification margin is expressed as a function of the network's Jacobian matrix. 
	
	\item Our approach includes a large class of DNNs. For example, we consider DNNs with the softmax layer at the network output; DNNs with various non-linearities such as the Rectified Linear Unit (ReLU), the sigmoid and the hyperbolic tangent; DNNs with pooling, such as down-sampling, average pooling and max-pooling; and networks with shortcut connections such as Residual Networks \cite{He2015}.

	\item Our analysis shows that the GE of a DNN can be bounded independently of its depth or width provided that the spectral norm of the Jacobian matrix in the neighbourhood of the training samples is bounded. We argue that this result gives a justification for a low GE of DNNs in practice. Moreover, it also provides an explanation for why training with the recently proposed weight normalization or batch normalization can lead to a small GE. In such networks the $\ell_2$-norm of the weight matrices is fixed and $\ell_2$-norm regularization does not apply. The analysis also leads to a novel Jacobian matrix-based regularizer, which can be applied to weight normalized or batch normalized networks.

	\item We provide a series of examples on the MNIST, CIFAR-10, LaRED and ImageNet datasets that validate our analysis and demonstrate the effectiveness of the Jacobian regularizer.
\end{itemize}

Our contributions differ from the existing works in many ways. In particular, the GE of DNNs has been studied via the algorithmic robustness framework in \cite{Xu2012a}. Their bounds are based on the per-unit $\ell_1$-norm of the weight matrices, and the studied loss is not relevant for classification. Our analysis is much broader, as it aims at bounding the GE of 0-1 loss directly and also considers DNNs with pooling. Moreover, our bounds are a function of the network's Jacobian matrix and are tighter than the bounds based on the norms of the weight matrices. 

The work in \cite{Huang2015} shows that learning transformations that are locally isometric is robust and leads to a small GE. Though they apply the proposed technique to DNNs they do not show how the DNN architecture  affects the GE as our work does.

The authors in \cite{An2015a} have observed that contractive DNNs with ReLUs trained with the hinge loss lead to a large classification margin. However, they do not provide any GE bounds. Moreover, their results are limited to DNNs with ReLUs, whereas our analysis holds for arbitrary non-linearities, DNNs with pooling and DNNs with the softmax layer.

The work in \cite{Rifai2011} is related to ours in the sense that it proposes to regularize auto-encoders by constraining the Frobenious norm of the encoder's Jacobian matrix. However, their work is more empirical and is less concerned with the classification margin or GE bounds. They use the Jacobian matrix to regularize the encoder whereas we use the Jacobian matrix to regularize the entire DNN.

Finally, our DNN analysis, which is based on the network's Jacobian matrix, is also related to the concept of sensitivity analysis that has been applied to feature selection for SVM and  neural networks \cite{Shen2008,Yang2008}, and for the construction of radial basis function networks \cite{Shi2005}, since the spectral norm of the Jacobian matrix quantifies the sensitivity of DNN output with respect to the input perturbation.

\subsection{Paper organization} 

Section~\ref{sec:problem_statement} introduces the problem of generalization error, including elements of the algorithmic robustness framework, and introduces DNN classifiers. Properties of DNNs are described in Section \ref{sec:DNNproperties}. The bounds on the classification margin of DNNs and their implication for the GE of DNNs are discussed in Section \ref{sec:margin_bounds}. Generalizations of our results are discussed in Section~\ref{sec:discussion}. Section~\ref{sec:experiments} presents experimental results. The paper is concluded in Section~\ref{sec:conclusions}. The proofs are deferred to the Appendix.

\subsection{Notation}

We use the following notation in the sequel: matrices, column vectors, scalars and sets are denoted by boldface upper-case letters ($\bX$), boldface lower-case letters ($\bx$), italic letters ($x$) and calligraphic upper-case letters ($\sX$), respectively. The convex hull of $\sX$ is denoted by $\text{conv}(\sX)$. $\bI_{N} \in \Rnn$ denotes the identity matrix, $\bZero_{M\times N} \in \Rmn$ denotes the zero matrix and $\bone_N \in \Ro{N}$ denotes the vector of ones. The subscripts are omitted when the dimensions are clear from the context. $\be_{k}$ denotes the $k$-th basis vector of the standard basis in $\Rn$. $\| \bx \|_2$ denotes the Euclidean norm of $\bx$, $\| \bX \|_{2}$ denotes the spectral norm of $\bX$, and $\| \bX \|_F$ denotes the Frobenious norm of $\bX$. The $i$-th element of the vector $\bx$ is denoted by $(\bx)_i$, and the element of the $i$-th row and $j$-th column of $\bX$ is denoted by $(\bX)_{ij}$. The covering number of $\sX$ with $d$-metric balls of radius $\rho$ is denoted by $\N(\sX;d,\rho)$. 

\section{Problem Statement} \label{sec:problem_statement}

We start by describing the GE in the framework of statistical learning. Then, we dwell on the GE bounds based on the robustness framework by Xu and Manor \cite{Xu2012a}. Finally, we present the DNN architectures studied in this paper.

\subsection{The Classification Problem and Its GE} \label{sec:GE}

We consider a classification problem, where we observe a vector $\bx \in \sX \subseteq \Ro{N}$ that has a corresponding class label $y \in \sY$. The set $\sX$ is called the input space, $\sY = \{1,2,\ldots, N_{\sY}\}$ is called the label space and $N_{\sY}$ denotes the number of classes. The samples space is denoted by $\sS = \sX \times \sY$ and an element of $\sS$ is denoted by $s = (\bx, y)$. We assume that samples from $\sS$ are drawn according to a probability distribution $P$ defined on $\sS$. A training set of $m$ samples drawn from $P$ is denoted by $S_m =\{ s_i \}_{i=1}^m =  \{(\bx_i, y_i) \}_{i=1}^m$. The goal of learning is to leverage the training set $S_m$ to find a classifier $g(\bx)$ that provides a label estimate $\hat{y}$ given the input vector $\bx$. In this work the classifier is a DNN, which is described in detail in Section~\ref{sec:DNN}. 

The quality of the classifier output is measured by the loss function $\ell( g(\bx), y)$, which measures the discrepancy between the true label $y$ and the estimated label $\hat{y} = g(\bx)$ provided by the classifier. Here we take the loss to be the 0-1 indicator function. Other losses such as the hinge loss or the categorical cross entropy loss are possible. The  empirical loss of the classifier $g(\bx)$ associated with the training set and the expected loss of the classifier $g(\bx)$ are defined as
\begin{IEEEeqnarray}{rCl}
	\ell_{\text{emp}}(g) = 1/m \sum_{s_i \in S_m} \ell \left(g(\bx_i), y_i \right)
\end{IEEEeqnarray}
and
\begin{IEEEeqnarray}{rCl}
	\ell_{\text{exp}}(g) = \E_{s \sim P} \left[ \ell\left( g(\bx), y \right) \right],
\end{IEEEeqnarray} 
respectively. An important question, which occupies us throughout this work, is how well $l_{\text{emp}}(g)$ predicts $l_{\text{exp}}(g)$.  The measure we use for quantifying the prediction quality is the difference between $l_{\text{exp}}(g)$ and $l_{\text{emp}}(g)$, which is called the \textit{generalization error}:
\begin{IEEEeqnarray}{rCl}
	\text{GE}(g) = |\ell_{\text{exp}}(g) - \ell_{\text{emp}}(g)| \,.
\end{IEEEeqnarray}

\subsection{The Algorithmic Robustness Framework}

In order to provide bounds to the GE for DNN classifiers we leverage the robustness framework \cite{Xu2012a}, which is described next.

The algorithmic robustness framework provides bounds for the GE based on the robustness of a learning algorithm that learns a classifier $g$ leveraging the training set $S_m$:
\begin{definition}[\cite{Xu2012a}] \label{def:robustness} Let $S_m$ be a training set and $\sS$ the sample space. A learning algorithm is $(K, \epsilon(S_m))$-robust if the sample space $\sS$ can be partitioned into $K$ disjoint sets denoted by $\sK_k$, $k = 1, \ldots, K$,
\begin{IEEEeqnarray}{l}
 \sK_k \subseteq \sS, \quad k = 1, \ldots, K, \\
  \sS = \cup_{k=1}^K \sK_k, \\
 \sK_k \cap \sK_{k'} = \emptyset, \quad\forall k \neq k',
\end{IEEEeqnarray}
such that for all $s_i \in S_m$ and all $s \in \sS$
\vspace{-0.15cm}
\begin{IEEEeqnarray}{rCl}
	s_i = (\bx_i, y_i) \in \sK_k \land s = (\bx, y)  \in \sK_k  \implies  \nonumber \\
	 | \ell(g(\bx_i), y_i) - \ell( g(\bx), y) | \leq \epsilon(S_m) \,. \label{eq:robustness_implication}
\end{IEEEeqnarray}
\end{definition}
Note that $s_i$ is an element of the training set and $s$ is an arbitrary element of the sample space $\sS$. Therefore, a robust learning algorithm chooses a classifier $g$ for which the losses of any $s$ and $s_i$ in the same partition $\sK_k$ are close. The following theorem provides the GE bound for robust algorithms.\footnote{Additional variants of this theorem are provided in \cite{Xu2012a}. 
}
\begin{theorem}[Theorem 3 in \cite{Xu2012a}] \label{th:robustness_generaliaztion}
	If a learning algorithm is $(K, \epsilon(S_m))$-robust and $\ell(g(\bx), y) \leq M$ for all $s = (\bx, y) \in \sS$, then for any $\delta > 0$, with probability at least $1- \delta$, 
	\begin{IEEEeqnarray}{rCl}
		\text{GE}(g)\leq \epsilon(S_m) + M \sqrt{\frac{2 K \log(2) + 2 \log(1/\delta)}{ m}} \,.  \label{eq:GE_bound}
	\end{IEEEeqnarray}
\end{theorem}
The first term in the GE bound in \eqref{eq:GE_bound} is constant and depends on the training set $S_m$. The second term behaves as $\mathcal{O}(1/\sqrt{m})$ and vanishes as the size of the training set $S_m$ approaches infinity. $M=1$ in the case of 0-1 loss, and $K$ corresponds to the number of partitions of the samples space $\sS$.

A bound on the number of partitions $K$ can be found by the covering number of the samples space $\sS$. The covering number is the smallest number of (pseudo-)metric balls of radius $\rho$ needed to cover $\sS$, and it is denoted by $\N(\sS;d,\rho)$, where $d$ denotes the (pseudo-)metric.\footnote{Note that we can always obtain a set of disjoint partitions from the set of metric balls used to construct the covering.} The space $\sS$ is the Cartesian product of a continuous input space $\sX$ and a discrete label space $\sY$, and we can write $\N(\sS;d,\rho) \leq N_{\sY} \cdot \N(\sX;d,\rho)$, where $N_\sY$ corresponds to the number of classes. The choice of metric $d$ determines how efficiently one may cover $\sX$. A common choice is the Euclidean metric
\begin{IEEEeqnarray}{rCl}
	d(\bx, \bx') = \| \bx - \bx' \|_2,,  \quad\bx, \bx' \in \sX \,,
\end{IEEEeqnarray}
which we also use in this paper. The covering number of many structured low-dimensional data models can be bounded in terms of their ``intrinsic'' properties, for example:
\begin{itemize}
	\item a Gaussian mixture model (GMM) with $L$ Gaussians and covariance matrices of rank at most $k$ leads to a covering number $\N(\sX;d,\rho) = L(1+ 2 / \rho)^k$  \cite{Mendelson2008};
	\item $k$-sparse signals in a dictionary with $L$ atoms have a covering number $\N(\sX;d,\rho) = {{L}\choose{k}} \left(1+ 2/\rho\right)^k$ \cite{Giryes2015};
	\item $C_M$ regular $k$-dimensional manifold, where $C_M$ is a constant that captures its ``intrinsic'' properties, has a covering number $\N(\sX;d,\rho) = \left( \frac{C_M}{\rho} \right)^k$ \cite{Verma2013}.
\end{itemize}

\subsubsection{Large Margin Classifier}
An example of a robust learning algorithm is the large margin classifiers, which we consider in this work. The classification margin is defined as follows:
\begin{definition}[Classification margin] 
The classification margin of a training sample $s_i = (\bx_i,y_i)$ measured by a metric $d$ is defined as 
\vspace{-0.15cm}
\begin{IEEEeqnarray}{rCl}
	\gamma^d(s_i) = \sup \{ a : \,d(\bx_i, \bx) \leq a  \implies g(\bx) = y_i \, \forall \bx\} \,. \label{eq:input_margin}
\end{IEEEeqnarray}
\end{definition}
The classification margin of a training sample $s_i$ is the radius of the largest metric ball (induced by $d$) in $\sX$ centered at $\bx_i$ that is contained in the decision region associated with class label $y_i$. The robustness of large margin classifiers is given by the following Theorem.
\begin{theorem}[Adapted from Example 9 in \cite{Xu2012a}] \label{th:max_margin_robust} If there exists $\gamma$ such that 
	\begin{IEEEeqnarray}{rCl}
		\gamma^d(s_i) > \gamma > 0  \quad \forall s_i \in S_m \,,
	\end{IEEEeqnarray}
	then the classifier $g(\bx)$ is $(N_{\sY} \cdot \N(\sX; d, \gamma/2), 0)$-robust. 
\end{theorem}
Theorems \ref{th:robustness_generaliaztion} and \ref{th:max_margin_robust} imply that the GE of a classifier with margin $\gamma$ is upper  bounded by (neglecting the $\log(1/\delta)$ term in \eqref{eq:GE_bound})
\begin{IEEEeqnarray}{rCl}
	\text{GE}(g) \lesssim \oneo{\sqrt{m}} \sqrt{ 2 \log(2) \cdot N_{\sY} \cdot \N(\sX;d,\gamma/2) } \,. \label{eq:GEform}
\end{IEEEeqnarray}
Note that in case of a large margin classifier the constant $\epsilon(S_m)$ in \eqref{eq:GE_bound} is equal to 0, and the GE approaches zero at a rate $\sqrt{m}$ as the number of training samples grows. The GE also increases sub-linearly with the number of classes $N_{\sY}$. Finally, the GE depends on the complexity of the input space $\sX$ and the classification margin via the covering number $\N(\sX;d,\gamma/2)$. 

For example, if we take  $\sX$ to be a $C_M$ regular $k$-dimensional manifold then the upper bound to the GE behaves as:
\begin{corollary} \label{th:GEmanifold}
Assume that $\sX$ is a (subset of) $C_M$ regular $k$-dimensional manifold, where $\N(\sX;d,\rho) \leq \left( \frac{C_M}{\rho} \right)^k$. Assume also that classifier $g(\bx)$ achieves a classification margin $\gamma$ and take $\ell(g(\bx_i), y_i)$ to be the 0-1 loss. Then for any $\delta > 0$, with probability at least $1- \delta$,
	\begin{IEEEeqnarray}{rCl}
		\text{GE}(g) &\leq& 
 \sqrt{\frac{\log(2) \cdot N_\sY \cdot 2^{k+1} \cdot (C_M)^k}{\gamma^k m}	} + \sqrt{\frac{2 \log(1/\delta)}{m}}  \,. \nonumber \\ \* \label{eq:GEbound_manifold}
	\end{IEEEeqnarray}
\begin{IEEEproof}
The proof follows directly from Theorems \ref{th:robustness_generaliaztion} and \ref{th:max_margin_robust}.
\end{IEEEproof}
\end{corollary}
Note that the role of the classifier is captured via the achieved classification margin $\gamma$. If we can always ensure a classification margin $\gamma = 1$, then the GE bound only depends on the dimension of the manifold $k$ and the manifold constant $C_M$. We relate this bound, in the context of DNNs, to other bounds in the literature in Section~\ref{sec:margin_bounds}.

\subsection{Deep Neural Network Classifier} \label{sec:DNN}

\begin{figure*}[t]
\centering
\begin{tikzpicture}[auto, thick, node distance=2cm, >=triangle 45,
	block/.style={
		draw,
		rectangle,
		minimum height = 0.7cm,
		align=center}]

\draw
	node at (0,0) []  (x) {$\mathbf{x}$}
	node [right of=x, block,node distance = 2.5cm] (f1) {$\phi_1(\bx,\theta_1)$}
	node [right of=f1, block,node distance = 3cm] (f2) {$\phi_2(\bz^1,\theta_2)$}
	node [right of=f2, block,,node distance = 3.5cm] (fL) {$\phi_L(\bz^{L-1},\theta_L)$}		
	node [right of=fL, rectangle,node distance = 2.5cm] (out) {$\bz$};
	
	\draw[->] (x) -- (f1);
	\draw[->] (f1) --  node {$\bz^1$}(f2);
	\draw[->,dashed] (f2) -- (fL);
	\draw[->] (fL) --  (out);	
\end{tikzpicture}
\caption{DNN transforms the input vector $\bx$ to the feature vector $\bz$ by a series of (non-linear) transforms.}
\label{fig:dnn}
\vspace{-0.15cm}
\end{figure*}
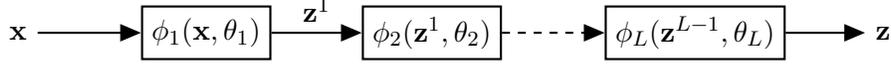

The DNN classifier is defined as 
\begin{IEEEeqnarray}{rCl}
	g(\bx) = \argmax_{i \in [N_\sY]} (f(\bx))_i \,, \label{eq:classifier}
\end{IEEEeqnarray}
where $(f(\bx))_i$ is the $i$-th element of the $N_\sY$ dimensional output of a DNN $f: \Ro{N} \to \Ro{N_\sY}$. We assume that $f(\bx)$ is composed of $L$ layers:
\begin{IEEEeqnarray}{rCl}
	f(\bx) = \phi_L( \phi_{L-1}( \cdots \phi_1(\bx, \theta_1), \cdots \theta_{L-1}), \theta_L) \,,
\end{IEEEeqnarray}
where $\phi_l(\cdot, \theta_l )$ represents the $l$-th layer with parameters $\theta_l$, $l = 1, \ldots, L$.  The output of the $l$-th layer is denoted by $\bz^l$, i.e. $\bz^l = \phi_l(\bz^{l-1}, \theta_l)$, $\bz^l \in \Ro{M_l}$; the input layer corresponds to $\bz^0 = \bx$; and the output of the last layer is denoted  by $\bz = f(\bx)$. Such a DNN is visualized in Fig.~\ref{fig:dnn}.
Next, we define various layers $\phi_l(\cdot, \theta_l)$ that are used in the modern state-of-the-art DNNs.
\subsubsection{Linear and Softmax Layers} \label{sec:softmax}
We start by describing the last layer of a DNN that maps the output of previous layer into $\Ro{N_\sY}$, where $N_\sY$ corresponds to the number of classes.\footnote{Assuming that there are $N_\sY$ one-vs.-all classifiers.} This layer can be linear:
\begin{IEEEeqnarray}{rCl}
	\bz = \hat{\bz}, \quad \hat{\bz} = \bW_L \bz^{L-1} + \bb_L, \label{eq:linear_layer}
\end{IEEEeqnarray}
where $\bW_L \in \Rt{N_\sY}{M_{L-1}}$ is the weight matrix associated with the last layer and $\bb \in \Ro{N_\sY}$ is the bias vector associated with the last layer. Note that according to \eqref{eq:classifier}, the $i$-th row of $\bW_L$ can be interpreted as a normal to  the hyperplane that separates class $i$ from the others. If the last layer is linear the usual choice of learning objective is the hinge loss. 

A more common choice for the last layer is the softmax layer:
\begin{IEEEeqnarray}{rCl}
	\bz = \zeta(\hat{\bz}) = e^{\hat{\bz}}/ \left( \bone^T e^{\hat{\bz}} \right) , \quad \hat{\bz} = \bW_L \bz^{L-1} + \bb_L  \,, \label{eq:softmax_layer}
\end{IEEEeqnarray}
where $\zeta(\cdot)$ is the softmax function and $\bW_L$ and $\bb_L$ are the same as in \eqref{eq:linear_layer}. Note that the exponential is applied element-wise. The elements of $\bz$ are in range $(0,1)$ and are often interpreted as ``probabilites'' associated with the corresponding class labels. The decision boundary between class $y_1$ and class $y_2$ corresponds to the hyperplane $\{\bz: (\bz)_{y_1} = (\bz)_{y_2} \}$. The softmax layer is usually coupled with categorical cross-entropy training objective.

For the remainder of this work we will take the softmax layer as the last layer of DNN, but note that all results still apply if the linear layer is used.

\subsubsection{Non-linear layers} \label{sec:FCDNN}

A non-linear layer is defined as
\begin{IEEEeqnarray}{rCl}
	\bz^{l}  = [\hat{\bz}^l ]_\sigma= [\bW_l \bz^{l-1} + \bb_l ]_\sigma \,, \label{eq:DNNlayer}
\end{IEEEeqnarray}
where $[\hat{\bz}^l]_\sigma$ represents the element-wise non-linearity applied to each element of  $\hat{\bz}^l \in \Ro{M_l}$, and $\hat{\bz}^l$ represents the linear transformation of the layer input: $\hat{\bz}^l = \bW_l \bz^{l-1} + \bb_l$. $\bW_l \in \Rt{M_{l}}{M_{l-1}}$ is the weight matrix and $\bb_l \in \Ro{M_l}$ is the bias vector.  The typical non-linearities are the ReLU, the sigmoid and the hyperbolic tangent. They are listed in Table~\ref{tab:nonlinearities}. The choice of non-linearity $\sigma$ is usually the same for all the layers in the network.
\begin{table*}[t]
\small
\center
\caption{Point-wise non-linearities} \label{tab:nonlinearities}
\begin{tabular}{cccc}
    \toprule
 Name & Function: $\sigma(x)$ & Derivative: $\frac{d}{dx}\sigma(x)$ & Derivative bound: $\sup_x \left|\frac{d }{dx}\sigma(x) \right|$ \\
 \midrule
 ReLU & $\max(x,0)$  & $\{1 \text{ if } x > 0; 0 \text{ if } x \leq 0\}$ & $\leq 1$ \\
 Sigmoid& $\frac{1}{1 + e^{-x}}$ & $\sigma(x)(1-\sigma(x)) = \frac{e^{-x}}{(1 + e^{-x})^2}$  & $\leq \oneo{4}$  \\
 Hyperbolic tangent & $\tanh(x) = \frac{e^x - e^{-x}}{e^x + e^{-x}}$ & $1 - \sigma(x)^2$ & $\leq 1$  \\
 \bottomrule
\end{tabular} 
\vspace{-0.15cm}
\end{table*}

Note that the non-linear layer in \eqref{eq:DNNlayer} includes the convolutional layers which are used in the convolutional neural networks. In that case the weight matrix is block-cyclic.

\subsubsection{Pooling layers} A pooling layer reduces the dimension of intermediate representation and is defined as
	\begin{IEEEeqnarray}{rCl}
		\bz^l = \bP^l(\bz^{l-1}) \bz^{l-1} \,, \label{eq:pooling_layer}
	 \end{IEEEeqnarray} 
where $\bP^l(\bz^{l-1})$ is the pooling matrix. The usual choices of pooling are down-sampling, max-pooling and average pooling. We denote by $\bp^l_i(\bz^{l-1})$ the $i$-th row of $\bP^l(\bz^{l-1})$ and assume that there are $M_{l}$ pooling regions $\sP_i$, $i = 1, \ldots, M_l$. In the case of down-sampling $\bp^l_i(\bz^{l-1}) = \be_{\sP_i(1)}$, where $\sP_i(1)$ is the first element of the pooling region $\sP_i$; in the case of max-pooling  $\bp^l_i(\bz^{l-1}) = \be_{j^\star}$, where $j^\star = \argmax_{j' \in \sP_i} |(\bz^{l-1})_{j'} |$; and in the case of average pooling $\bp^l_i(\bz^{l-1}) = \oneo{| \sP_i |} \sum_{j \in \sP_i} \be_j$.

\section{The Geometrical Properties of Deep Neural Networks} \label{sec:DNNproperties}

\begin{figure}[t]
	\centering
	\subfigure[Input space.]{\includegraphics[height=0.3\textwidth]{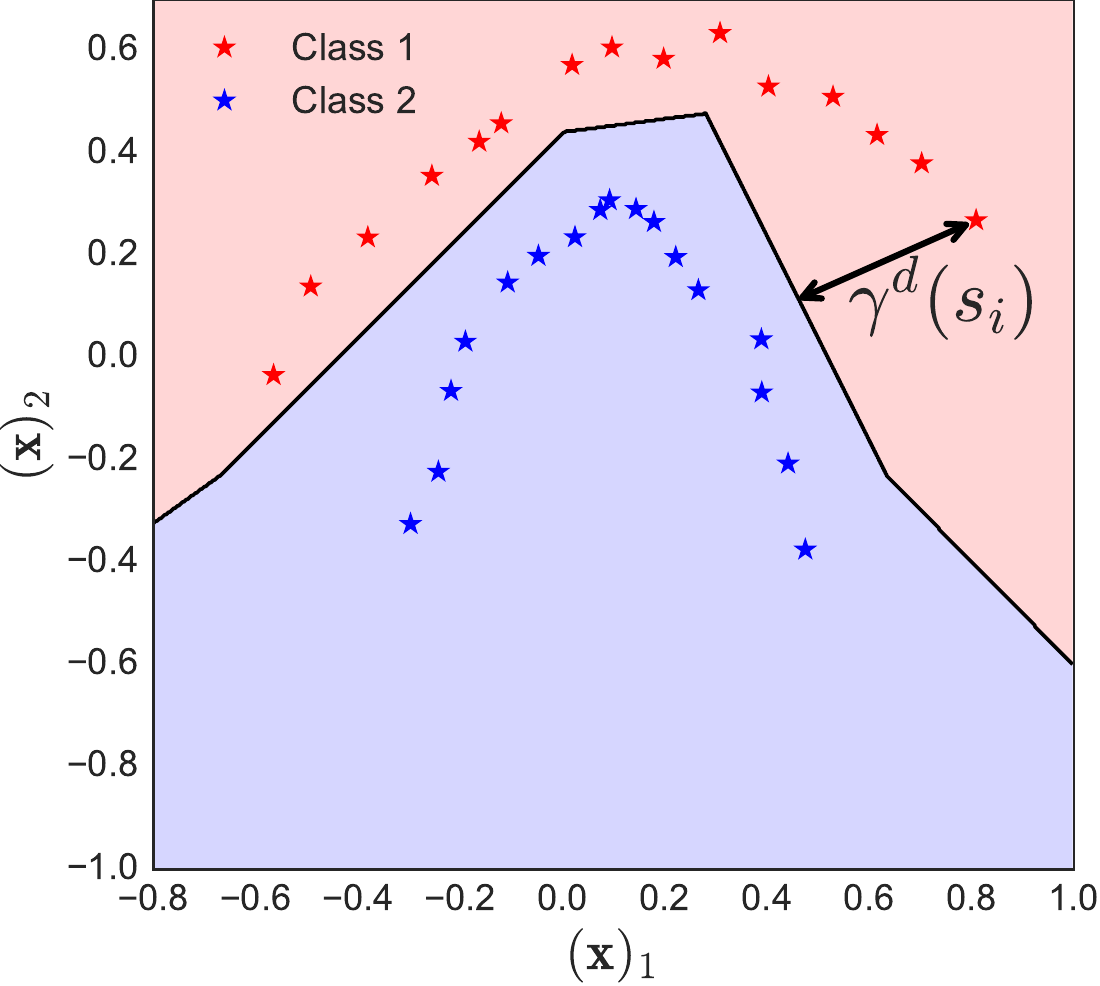}} 
	\subfigure[Output space.]{\includegraphics[height=0.3\textwidth]{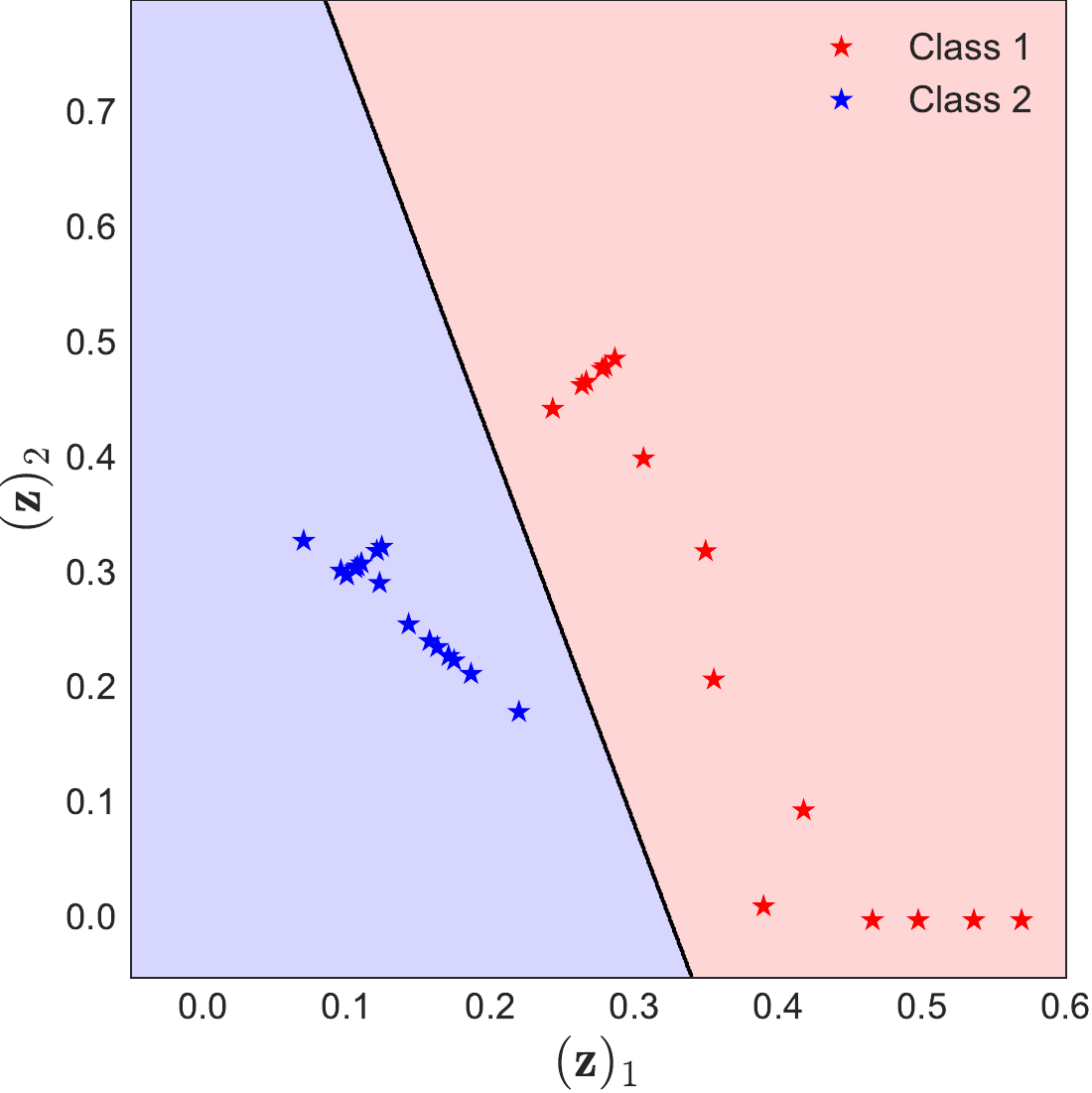}} 
\caption{Decision boundaries in the input space and in the output space. Plot (a) shows samples of class 1 and 2 and the decision regions produced by a two-layer network projected into the input space. Plot (b) shows the samples transformed by the network and the corresponding decision boundary at the network output.} \label{fig:network}	
\vspace{-0.15cm}
\end{figure}

The classification margin introduced in Section \ref{sec:GE} is a function of the decision boundary in the input space. This is visualized in Fig.~\ref{fig:network} (a). However, a training algorithm usually optimizes the decision boundary at the network output (Fig.~\ref{fig:network} (b)), which does not necessarily imply a large classification margin. In this section we introduce a general approach that allows us to bound the expansion of distances between the network input and its output. In Section~\ref{sec:margin_bounds} we use this to establish bounds of the classification margin and the GE bounds that are independent of the network depth or width.

We start by defining the Jacobian matrix (JM) of the DNN $f(\bx)$:
\begin{IEEEeqnarray}{rCl}
	\bJ(\bx) =  \frac{d f(\bx)} {d \bx} =  \prod_{l=1}^L \frac{d \phi_l(\bz^{l-1})}{d \bz^{l-1}} \cdot \frac{d \phi_1(\bx)}{d \bx}. \label{eq:JM}
\end{IEEEeqnarray}
Note that by the properties of the chain rule, the JM is computed as the product of the JMs of the individual network layers, evaluated at the appropriate values of the layer inputs $\bx, \bz^1, \ldots, \bz^{L-1}$. We use the JM to establish a relation between a pair of vectors in the input space and the output space.
\begin{theorem} \label{th:jacobian_difference}
	For any $\bx, \bx' \in \sX$ and a DNN $f(\cdot)$, we have
	\begin{IEEEeqnarray}{rCl}
	f(\bx')  - f(\bx) &=& \int_{0}^1 \bJ(\bx + t (\bx' - \bx))  \,dt \, (\bx' -  \bx) \label{eq:jacobian_integral}  \\
		&=& \bJ_{\bx, \bx'} (\bx' - \bx),
\end{IEEEeqnarray}
where 
\begin{IEEEeqnarray}{rCl}
\bJ_{\bx, \bx'} = \int_{0}^1 \bJ(\bx + t (\bx' - \bx))  \,dt \label{eq:average_jacobian}
\end{IEEEeqnarray}
is the average Jacobian on the line segment between $\bx$ and $\bx'$.
\begin{IEEEproof} The proof appears in Appendix~\ref{sec:th:proof:jacobian_difference}.%
\end{IEEEproof}
\end{theorem}
As a direct consequence of Theorem \ref{th:jacobian_difference} we can bound the distance expansion between $\bx$ and $\bx'$ at the output of the network $f(\cdot)$:
\begin{corollary} \label{th:dist_inequality}
For any $\bx, \bx' \in \sX$ and a DNN $f(\cdot)$, we have
\begin{IEEEeqnarray}{rCl}
 \| f(\bx') - f(\bx) \|_2 &=& \|\bJ_{\bx, \bx'} (\bx' - \bx) \|_2  \nonumber \\\
 				&\leq& \sup_{\bx'' \in \text{conv}(\sX)}\|\bJ(\bx'') \|_2 \| \bx' -\bx \|_2 \,.  \label{eq:euc_dist_inequality}
\end{IEEEeqnarray}
\begin{IEEEproof}
The proof appears in Appendix~\ref{sec:th:proof:dist_inequality}.
\end{IEEEproof}
\end{corollary}
Note that we have established that $\bJ_{\bx, \bx'}$ corresponds to a linear operator that maps the vector $\bx' - \bx$ to the vector $f(\bx') - f(\bx)$. This implies that the maximum distance expansion of the network $f(\bx)$ is bounded by the maximum spectral norm of the network's JM. Moreover, the JM of $f(\bx)$ corresponds to the product of JMs of all the layers of $f(\bx)$ as shown in \eqref{eq:JM}. It is possible to calculate the JMs of all the layers defined in Section~\ref{sec:DNN}:

\subsubsection{Jacobian Matrix of Linear and Softmax Layers}
The JM of the linear layer defined in \eqref{eq:linear_layer} is equal to the weight matrix
\begin{IEEEeqnarray}{rCl}
	\frac{d \bz}{d \bz^{L-1}} = \bW_L	 \,.
\end{IEEEeqnarray}
Similarly, in the case of softmax layer defined in \eqref{eq:softmax_layer} the JM is
\begin{IEEEeqnarray}{rCl}
	\frac{d \bz}{d \bz^{L-1}} &=&	\frac{d \bz}{d \hat{\bz}} \cdot 	\frac{d \hat{\bz}}{d \bz^{L-1}} \nonumber \\ 
	&=& \left( - \zeta(\hat{\bz}) \zeta(\hat{\bz})^T + \diag(\zeta(\hat{\bz}) \right) \cdot \bW_L \,.
\end{IEEEeqnarray}
Note that $\left( - \zeta(\hat{\bz}) \zeta(\hat{\bz})^T + \diag(\zeta(\hat{\bz}) \right)$ corresponds to the JM of the softmax function $\zeta(\hat{\bz})$.
\subsubsection{Jacobian Matrix of Non-Linear Layers}

The JM of the non-linear layer \eqref{eq:DNNlayer} can be derived in the same way as the JM of the softmax layer. We first define the JM of the point-wise non-linearity, which is a diagonal matrix\footnote{Note that in case of  ReLU the derivative of $max(x, 0)$ is not defined for $x = 0$, and we need to use subderivatives (or subgradients) to define the JM. We avoid this technical complication and simply take the derivative of
$max(x, 0)$ to be 0 when $x = 0$. Note that this does not change the results in any way because the subset of $\sX$ for which the derivatives are not defined has zero measure.}
\begin{IEEEeqnarray}{rCl}
	\left(\frac{d \bz^l}{d \hat{\bz}^{l}} \right)_{ii} = \frac{d\sigma \left( (\hat{\bz}^l)_{i} \right)}{d (\hat{\bz}^l)_i}, \quad i  = 1, \ldots, M_l \,. \label{eq:JM_nonlinear_diag}
\end{IEEEeqnarray}
The derivatives associated with various non-linearities are provided in Table~\ref{tab:nonlinearities}.
The JM of the non-linear layer can be expressed as
\begin{IEEEeqnarray}{rCl}
	\frac{d \bz^l}{d \hat{\bz}^{l-1}} = \frac{d \bz^l}{d \hat{\bz}^{l}} \cdot \bW_l \,. \label{eq:JMFC}
\end{IEEEeqnarray}

\subsubsection{Jacobian Matrix of Pooling Layers}
The pooling operator defined in \eqref{eq:pooling_layer} is linear or a piece-wise linear operator. The corresponding JM is therefore also linear or piece-wise linear and is equal to:
\begin{IEEEeqnarray}{rCl}
	\bP^l(\bz^{l-1}) \,.  \label{eq:JMpool}
\end{IEEEeqnarray}

The following Lemma collects the bounds on the spectral norm of the JMs for all the layers defined in Section~\ref{sec:DNN}. 
\begin{lemma} \label{th:jacobian_spectral_norm_bound}
The following statements hold:
\begin{enumerate}
	\item The spectral norm of JMs of the linear layer in \eqref{eq:linear_layer}, the softmax layer in \eqref{eq:softmax_layer} and non-linear layer in \eqref{eq:DNNlayer} with the ReLU, Sigmoid or Hyperbolic tangent non-linearities is upper bounded by
	\begin{IEEEeqnarray}{rCl}
		\left \| \frac{d \bz^l}{d \hat{\bz}^{l-1}}  \right \|_2 \leq \| \bW_l \|_2 \leq \| \bW_l \|_F \,.
	\end{IEEEeqnarray}
	
	\item Assume that the pooling regions of the down-sampling, max-pooling and average pooling operators are non-overlapping. Then the spectral norm of their JMs can be upper bounded by
	\begin{IEEEeqnarray}{rCl}
		\left \| \frac{d \bz^l}{d \hat{\bz}^{l-1}} \right \|_2 \leq 1 \,.
	\end{IEEEeqnarray}
\end{enumerate}
\begin{IEEEproof}
The proof appears in Appendix~\ref{sec:th:proof:jacobian_spectral_norm_bound}.
\end{IEEEproof}
\end{lemma}
Lemma \ref{th:jacobian_spectral_norm_bound} shows that the spectral norms of all layers can be bounded in terms of their weight matrices. As a consequence, the spectral norm of the JM is bounded by the product of the spectral norms of the weight matrices. We leverage this facts to provide GE bounds in the next section. 

We also briefly explore a relationship between the Jacobian matrix and the Fisher information matrix. To simplify the derivations we assume $M=1$, $N=1$, $\bx^\prime = \bx + \theta \bn$ and \mbox{$\bn \sim \N(0, 1)$}, where $\theta$ is the model parameter and $\bx$ is deterministic. The Fisher information $F(\theta)$ measures how much information about the  parameter $\theta$ is contained in the random variable $\by = f(\bx')$, where $f$ represents a DNN. In this particular case the Fisher information is given as
\begin{IEEEeqnarray}{rCl}
	F(\theta) &=& \E_\bn \left[ \left( \frac{d \log f(\bx')}{d \theta} \right)^2\right ] \nonumber  \\
	&=& \E_\bn \left[ \left( \frac{d \log f(\bx')}{d f(\bx')} \frac{d f(\bx')}{d \bx'} \frac{d \bx'}{\theta}  \right)^2\right ] \nonumber \\
	&=& \E_\bn \left[ \left( \frac{d \log f(\bx')}{d f(\bx')} \bJ(\bx') \bn \right)^2\right ]\,. \label{eq:Fin_jac}
\end{IEEEeqnarray}
In our setup the parameter $\theta$ can be interpreted as a magnitude of the input perturbation. It is clear from \eqref{eq:Fin_jac} that a small norm of the Jacobian matrix leads to a small Fisher information, which indicates that the distribution of $\by$ is not very informative about the parameters $\theta$. By ensuring that the norm of the Jacobian is small we then naturally endow the network with robustness against perturbations of the input.

\section{Generalization error of a Deep Neural Network Classifier} \label{sec:margin_bounds}

In this section we provide the classification margin bounds for DNN classifiers that allow us to bound the GE. We follow the common practice and assume that the networks are trained by a loss that promotes  separation of different classes at the network output, e.g. categorical cross entropy loss or the hinge loss. In other words, the training aims at maximizing the score of each training sample, where the score is defined as follows.
\begin{definition}[Score] Take score of a training sample \mbox{$s_i = (\bx_i, y_i)$}
\vspace{-0.15cm}
\begin{IEEEeqnarray}{rCl}
o(s_i) &=& \min_{j \neq y_i} \sqrt{2} (\boldsymbol{\delta}_{y_i} - \boldsymbol{\delta}_{j})^T  f(\bx_i)  \,, \label{eq:output_score}
\end{IEEEeqnarray}
where $\boldsymbol{\delta}_i \in \Ro{N_\sY}$ is the Kronecker delta vector with $(\boldsymbol{\delta}_i)_i = 1$. 
\end{definition}
Recall the definition of the classifier $g(\bx)$ in \eqref{eq:classifier} and note that the decision boundary between class $i$ and class $j$ in the feature space $\sZ$ is given by the hyperplane $\{\bz : (\bz)_i = (\bz_j) \}$. A positive score indicates that at the network output, classes are separated by a margin that corresponds to the score. However, a large score $o(s_i)$ does not necessarily imply a large classification margin $\gamma^d(s_i)$. Theorem \ref{th:margin_bound_euc} provides classification margin bounds expressed as a function of the score and the properties of the network.
\begin{theorem} \label{th:margin_bound_euc}
	Assume that a  DNN classifier $g(\bx)$, as defined in \eqref{eq:classifier}, classifies a training sample $\bx_i$ with the score $o(s_i) > 0$. Then the classification margin can be bounded as
	\begin{IEEEeqnarray}{rCl}
		\gamma^{d}(s_i) &\geq & \frac{o(s_i)}{\sup_{\bx: \|\bx - \bx_i\|_2 \leq \gamma^{d}(s_i)} \left \|  \bJ(\bx) \right \|_2} \triangleq \gamma^d_1(s_i) \label{eq:margin_bound_euc1}\\
					 & \geq & \frac{o(s_i)}{\sup_{\bx \in \text{conv}(\sX)}  \left \|  \bJ(\bx) \right \|_2}  \triangleq \gamma^d_2(s_i)	  \label{eq:margin_bound_euc2} \\
					& \geq &	 \frac{o(s_i)}{\prod_{\bW_l \in \sW}  \left \|  \bW_l \right \|_2}  \triangleq \gamma^d_3(s_i) \label{eq:margin_bound_euc3}\\
					& \geq &	 \frac{o(s_i)}{\prod_{\bW_l \in \sW}  \left \|  \bW_l \right \|_F}  \triangleq \gamma^d_4(s_i)	\label{eq:margin_bound_euc4}  \,,
\end{IEEEeqnarray}		
where $\sW$ is the set of all weight matrices of $f(\bx)$.
\begin{IEEEproof}
The proof appears in Appendix~\ref{sec:th:proof:margin_bound_euc}.
\end{IEEEproof}
\end{theorem}
Given the bounds of the classification margin we can specialize Corollary~\ref{th:GEmanifold} to DNN classifiers.
\begin{corollary} \label{th:GEmanifold_DNN}
Assume that $\sX$ is a (subset of) $C_M$ regular $k$-dimensional manifold, where $\N(\sX;d,\rho) \leq \left( \frac{C_M}{\rho} \right)^k$. Assume also that DNN classifier $g(\bx)$ achieves a lower bound to the classification margin $\gamma^d_b(s_i) > \gamma_b$ for $b \in \{ 1,2,3,4\}$, $\forall s_i \in S_m$ and take $\ell(g(\bx_i), y_i)$ to be the 0-1 loss. Then for any $\delta > 0$, with probability at least $1- \delta$,
	\begin{IEEEeqnarray}{rCl}
		\text{GE}(g) &\leq& 
 \sqrt{\frac{\log(2) \cdot N_\sY \cdot 2^{k+1} \cdot (C_M)^k}{\gamma_b^k m}	} + \sqrt{\frac{2 \log(1/\delta)}{m}}  \,. \nonumber \\ \* \label{eq:GEbound_manifold_DNN}
	\end{IEEEeqnarray}
\begin{IEEEproof}
	The proof follows from Theorems \ref{th:robustness_generaliaztion}, \ref{th:max_margin_robust} \mbox{and \ref{th:margin_bound_euc}}.
\end{IEEEproof}
\end{corollary}

Corollary 3 suggests that the GE will be bounded by $C \frac{1}{\sqrt{m}} \gamma^{-k/2}$, where $C = \sqrt{\log(2) \cdot N_\sY 2^{k+1} (C_M)^k}$, provided that the classification margin bounds satisfy $\gamma_b^d(s_i) > \gamma$ for some $b \in \{1,2,3,4\}, \forall s_i \in S_m$. 

We now leverage the classification margin bounds in Theorem~4 to construct constraint sets $\sW_b = \{\bW_l \in \sW : \gamma_b^d(s_i) > \gamma \forall s_i \}$, $b \in {1,2,3,4}$ such that  $\sW \in \sW_b$ ensures that the GE is bounded by $C \frac{1}{\sqrt{m}} \gamma^{-k/2}$. Using \eqref{eq:margin_bound_euc1}-\eqref{eq:margin_bound_euc4} we obtain
\begin{IEEEeqnarray}{rCl}
	\sW_1 &=& \big \{ \bW_l \in \sW : \sup_{\bx: \|\bx - \bx_i\|_2 \leq \gamma^{d}(s_i)} \left \|  \bJ(\bx) \right \|_2 < \gamma \cdot o(s_i) \big . \nonumber \\ && \quad \quad \quad \quad \quad \qquad \qquad \qquad \qquad \big . \forall s_i = (\bx_i,y_i)  \big \}, \label{eq:cset1} \\
	\sW_2 &=& \big \{ \bW_l \in \sW : \sup_{\bx \in \text{conv}(\sX)}  \left \|  \bJ(\bx) \right \|_2 < \gamma \cdot o(s_i) \big . \nonumber \\ && \quad \quad \quad \quad \quad \qquad \qquad \qquad \qquad \big . \forall s_i = (\bx_i,y_i)  \big \}, \label{eq:cset2} \\
	\sW_3 &=& \big \{ \bW_l \in \sW : \prod_{\bW_l \in \sW}  \left \|  \bW_l \right \|_2 < \gamma \cdot o(s_i) \big . \nonumber \\ && \quad \quad \quad \quad \quad \qquad \qquad \qquad \qquad \big . \forall s_i = (\bx_i,y_i)  \big \}, \label{eq:cset3} \\
	\sW_4 &=& \big \{ \bW_l \in \sW : \prod_{\bW_l \in \sW}  \left \|  \bW_l \right \|_F < \gamma \cdot o(s_i) \big . \nonumber \\ && \quad \quad \quad \quad \quad \qquad \qquad \qquad \qquad \big . \forall s_i = (\bx_i,y_i)  \big \}. \label{eq:cset4}
\end{IEEEeqnarray}	

Note that while we want to maximize the score $o(s_i)$, we also need to constrain the network's Jacobian matrix $\bJ(\bx)$ (following $\sW_1$ and $\sW_2$) or the weight matrices $\bW_l\in \sW$ (following $\sW_3$ and $\sW_4$). This stands in line with the common training rationale of DNN in which we do not only aim at maximizing the score of the training samples to ensure a correct classification of the training set, but also have a regularization that constrains the network parameters, where this combination eventually leads to a lower GE. The constraint sets in \eqref{eq:cset1}-\eqref{eq:cset4} impose different regularization techniques:
\begin{itemize}

%

\item The term $\sup_{\bx: \|\bx - \bx_i\|_2 \leq \gamma^{d}(s_i)} \left \|  \bJ(\bx) \right \|_2 < \gamma \cdot o(s_i)$ in \eqref{eq:cset1} considers only the supremum of the spectral norm of the Jacobian matrix evaluated at the points within $\N_i = \{ \bx :  \|\bx - \bx_i\|_2 \leq \gamma^{d}(s_i) \}$, where $\gamma^{d}(s_i)$ is the classification margin of training sample $s_i$ (see Definition 2). We can not compute the margin $\gamma^{d}(s_i)$, but can still obtain a rationale for regularization: as long as the spectral norm of the Jacobian matrix is bounded in the neighbourhood of a training sample $\bx_i$ given by $\N_i$ we will have the GE guarantees. 

\item The constraint on the Jacobian matrix $\sup_{\bx \in \text{conv}(\sX)}  \left \|  \bJ(\bx) \right \|_2 < \gamma \cdot o(s_i)$ in \eqref{eq:cset2} is more restrictive as it requires bounded spectral norm for all samples $\bx$ in the convex hull of the input space $\sX$.

\item The constraints in \eqref{eq:cset3} and \eqref{eq:cset4} are of similar form, $\prod_{\bW_l \in \sW}  \left \|  \bW_l \right \|_2 < \gamma \cdot o(s_i)$ and $\prod_{\bW_l \in \sW}  \left \|  \bW_l \right \|_F < \gamma \cdot o(s_i)$, respectively. Note that the weight decay, which aims at bounding the Frobenious norms of the weight matrices might be used to satisfy the constrains in \eqref{eq:cset4}. However, note also that the bound based on the spectral norm in \eqref{eq:cset3} is tighter than one based on the Frobenious norm in \eqref{eq:cset4}.  For example, take $\bW_l \in \sW$ to have orthonormal rows and be of dimension $M \times M$. Then the constraint in \eqref{eq:cset3}, which is based on the spectral norm, is of the form $1 < \gamma o(s_i)$ and the constraint in \eqref{eq:cset4}, which is based on the Frobenious norm, is $M^{L/2} < \gamma o(s_i)$. In the former case we have a constraint on the score, which is independent of the network width or depth. In the latter the constraint on the output score is exponential in network depth and polynomial in network width. The difference is that the Frobenious norm does not take into account the correlation (angles) between the rows of the weight matrix $\bW_l$, while the spectral norm does. Therefore, the bound based on the Frobenious norm corresponds to the worst case when all the rows of $\bW_l$ are aligned. In that case $\| \bW_l \|_F = \| \bW_l \|_2 = \sqrt{M}$. On the other hand, if the rows of $\bW_l$ are orthonormal $\| \bW_l \|_F = \sqrt{M}$, but $\| \bW_l \|_2 = 1$. 
\end{itemize}

\begin{remark}
 To put results into perspective we compare our GE bounds to the GE bounds based on the Rademacher complexity in  \cite{Neyshabur2015}, which hold for DNNs with ReLUs. The work in \cite{Neyshabur2015} shows that if
\begin{IEEEeqnarray}{rCl}
	\sW \in \sW_F = \{ \bW_i \in \sW : \prod_{i=1}^L \| \bW_i \|_F	< C_F \} \label{eq:cset5}
\end{IEEEeqnarray}
 and the energy of training samples is bounded then:
 \vspace{-0.15cm}
\begin{IEEEeqnarray}{rCl}
	GE(g) \lessapprox \oneo{\sqrt{m}} 2^{L-1}  C_F  \,. \label{eq:erc_bahaviour}
\end{IEEEeqnarray}
Although the bounds \eqref{eq:GEbound_manifold_DNN} and \eqref{eq:erc_bahaviour} are not directly comparable, since the bounds based on the robustness framework rely on an underlying assumption on the data (covering number), there is still a remarkable difference between them. The behaviour in \eqref{eq:erc_bahaviour} suggests that the GE grows exponentially with the network depth even if the product of the Frobenious norms of all the weight matrices is fixed, which is due to the term $2^L$. The bound in (34) and the constraint sets in  \eqref{eq:cset1}-\eqref{eq:cset4}, on the other hand, imply that the GE does not increase with the number of layers provided that the spectral/Frobenious norms of the weight matrices are bounded. Moreover, if we take the DNN to have weight matrices with orthonormal rows then the GE behaves as $\oneo{\sqrt{m}} (C_M)^{k/2}$ (assuming $o(s_i) \geq 1, i = 1, \ldots, m$), and therefore relies only on the complexity of the underlying data manifold and not on the network depth. This provides a possible answer to the open question of \cite{Neyshabur2015} that depth independent capacity control is possible in DNNs with ReLUs.
\end{remark}

\begin{remark} An important value of our bounds is that they provide an additional explanation to the success of state-of-the-art DNN training techniques such as batch normalization \cite{Ioffe2015} and eight normalization \cite{Salimans2016}. 

Weight normalized DNNs have weight matrices with normalized rows, i.e.
\begin{IEEEeqnarray}{rCl}
	\bW_l = \diag( \hat{\bW}_l^T \hat{\bW}_l )^{-1} \hat{\bW}_l  \,, \label{eq:weight_normalization}
\end{IEEEeqnarray}
where $\diag(\cdot)$ denotes the diagonal part of the matrix. While the main motivation for this method is a faster training, the authors also show empirically that such networks achieve good generalization. Note that for row-normalized weight matrices $\| \bW_l \|_F = \sqrt{M_l}$ and therefore the bounds based on the Frobenious norm can not explain the good generalization of such networks as adding layers or making $\bW_l$ larger will lead to a larger GE bound. However, our bound in (34) and the constraint sets in \eqref{eq:cset1}-\eqref{eq:cset3} show that a small Frobenious norm of the weight matrices is not crucial for a small GE. A supporting experiment is presented in Section~\ref{sec:experiment_weight_normalization}.

We also note that batch normalization also leads to row-normalized weight matrices in DNNs with ReLUs:\footnote{To simplify the derivation we omit the bias vectors and therefore also the centering applied by the batch normalization. This does not affect the generality of the result. We also follow \cite{Neyshabur2015a} and omit the batch normalization scaling, as it can be included into the weight matrix of the layer following the batch normalization. We also omit the regularization term and assume that the matrices are invertible.}
\end{remark}

\begin{theorem} \label{th:batchnorm}
Assume that the non-linear layers of a DNN with ReLUs are batch normalized as:
\begin{IEEEeqnarray}{rCl}
	\bz^{l+1} = [\bN \left(\{\bz_i^l\}_{i=1}^m, \bW_l \right)\, \hat{\bz^l}]_{\sigma}, \quad \hat{\bz}^{l} = \bW_l  \bz^{l} \,,
\end{IEEEeqnarray}
where $\sigma$ denotes the ReLU non-linearity and
\begin{IEEEeqnarray}{rCl}
\bN \left(\{\bz_i\}_{i=1}^m, \bW \right) = \diag \left( \sum_{i=1}^m \bW \bz_i \bz_i^T \bW^T  \right)^{-\oneo{2}} 
\end{IEEEeqnarray}
is the normalization matrix. Then all the weight matrices are row normalized. The exception is the weight matrix of the last layer, which is of the form $\bN(\{\bz_i^{L-1}\}_{i=1}^m, \bW_L) \bW_L$.
\begin{IEEEproof}
The proof appears in Appendix~\ref{sec:batchnorm}.
\end{IEEEproof}
\end{theorem}

\subsection{Jacobian Regularizer}

The constraint set \eqref{eq:cset1} suggests that we can regularize the DNN by bounding the norm of the network's JM for the inputs close to $\bx_i$. Therefore, we propose to penalize the norm of the network's JM evaluated at each training sample $\bx_i$,
\begin{IEEEeqnarray}{rCl}
	R_J(\sW) = \oneo{m} \sum_{i=1}^m \| \bJ(\bx_i) \|_2^2 \,.
\end{IEEEeqnarray}
The implementation of such regularizer requires computation of its gradients or subgradients. In this case the computation of the subgradient of the spectral norm requires the calculation of a SVD decomposition \cite{Watson1992}, which makes the proposed regularizer inefficient. To circumvent this, we propose a surrogate regularizer based on the Frobenious norm of the Jacobian matrix:
\begin{IEEEeqnarray}{rCl}
	R_F(\sW) = \oneo{m} \sum_{i=1}^m \| \bJ(\bx_i) \|_F^2. \label{eq:regF}
\end{IEEEeqnarray}
Note that the Frobenious norm and the spectral norm are related as follows: $1/\text{rank}(\bJ(\bx_i)) \| \bJ(\bx_i) \|_F^2 \leq \| \bJ(\bx_i) \|_2^2 \leq \| \bJ(\bx_i) \|_F^2$, which justifies using the surrogate regularizer. We will refer to $R_F(\sW)$ as the Jacobian regularizer.

\subsubsection{Computation of Gradients and Efficient Implementation} Note that the $k$-th row of $\bJ(\bx_i)$ corresponds to the gradient of $(f(\bx))_k$ with respect to the input $\bx$ evaluated at $\bx_i$. It is denoted by $\bg_k(\bx_i) = \frac{d (f(\bx))_k} {d \bx} |_{\bx = \bx_i}$. Now we can write
\begin{IEEEeqnarray}{rCl}
	R_F(\sW) = \oneo{m} \sum_{i=1}^m \sum_{k=1}^M \bg_k(\bx_i)  \bg_k(\bx_i)^T.
\end{IEEEeqnarray}
As the regularizer will be minimized by a gradient descent algorithm we need to compute its gradient with respect to the DNN parameters. First, we express $\bg_k(\bx_i)$ as 
\begin{IEEEeqnarray}{rCl}
	\bg_k(\bx_i) = \bg_k^{l}(\bx_i) \bW_l \bJ^{l-1}(\bx_i)
\end{IEEEeqnarray}
where $\bg_k^{l}(\bx_i) = \frac{d (f(\bx))_k} {d \hat{\bz}^l} |_{\bx = \bx_i}$ is the gradient of $(f(\bx))_k$ with respect to $\hat{\bz}^l$ evaluated at the input $\bx_i$ and $\bJ^{l-1}(\bx_i) = \frac{d \bz^{l-1}} {d \bx} |_{\bx = \bx_i}$ is the JM of $l-1$-th layer output $\bz^{l-1}$ evaluated at the input $\bx_i$. The gradient of $\bg_k(\bx_i)  \bg_k(\bx_i)^T$ with respcet to $\bW_l$ is then given as \cite{petersen2008the-matrix}
\begin{IEEEeqnarray}{rCl}
\nabla_{\bW_l} \left(\bg_k(\bx_i)  \bg_k(\bx_i)^T\right) =2  \bg_k^{l}(\bx_i)^T \bg_k^{l}(\bx_i) \bW_l \bJ^{l-1}(\bx_i). \nonumber 
\end{IEEEeqnarray}
The computation of the gradient of the regularizer at layer $l$ requires the computation of gradients $\bg_k^{l}(\bx_i)$, $k=1,\ldots, M$, $i = 1, \ldots, m$ and the computation of the Jacobian matrices $\bJ^{l-1}(\bx_i)$, $i = 1, \ldots, m$. The computation of the gradient of a typical loss used for training DNN usually involves a computation of $m$ gradients with computational complexity similar to the computational complexity of $\bg_k^{l}(\bx_i)$. Therefore, the computation of gradients required for an implementation of the Jacobian regularizer can be very expensive.

To avoid excessive computational complexity we propose a simplified version of the regularizer \eqref{eq:regF}, which we name per-layer Jacobian regularizer. The per-layer Jacobian regularizer at layer $l$ is defined as 
\begin{IEEEeqnarray}{rCl}
	R_F^l(\bW_l) = \oneo{m} \sum_{i=1}^m \tilde{\bg}_{\pi(i)}^{l-1}(\bx_i)  (\tilde{\bg}_{\pi(i)}^{l-1}(\bx_i))^T, \label{eq:regFefficient}
\end{IEEEeqnarray}
where $\tilde{\bg}_{\pi(i)}^{l-1}(\bx_i) = \frac{d (f(\bx))_{\pi(i)}} {d \bz^{l-1}} |_{\bx = \bx_i}$, and $\pi(i) \in \{1,\ldots, M\}$ is a random index. Compared to \eqref{eq:regF} we have made two simplifications. First, we assumed that input of layer $l$ is fixed. This way we do not need to compute the JM $\bJ^{l-1}(\bx_i)$ between the output of the layer $l-1$ and the input. Second, by choosing only one index $\pi(i)$ per training sample we have to compute only one additional gradient per training sample. This significantly reduces the computational complexity. The gradient of $\tilde{\bg}_{\pi(i)}^{l-1}(\bx_i) (\tilde{\bg}_{\pi(i)}^{l-1}(\bx_i))^T$ is simply 
\begin{IEEEeqnarray}{rCl}
\nabla_{\bW_l} \left(\tilde{\bg}_{\pi(i)}^{l-1}(\bx_i) (\tilde{\bg}_{\pi(i)}^{l-1}(\bx_i))^T\right) =2  \bg_{\pi(i)}^{l}(\bx_i)^T \bg_{\pi(i)}^{l}(\bx_i) \bW_l. \nonumber 
\end{IEEEeqnarray}

We demonstrate the effectiveness of this regularizers in Section~\ref{sec:experiments}.

\section{Discussion} \label{sec:discussion}

In the preceding sections we analysed the standard feed-forward DNNs and their classification margin measured in the Euclidean norm. We now briefly discuss how our results extend to other DNN architectures and different margin metrics.

\subsection{Beyond Feed Forward DNN}

There are various DNN architectures such as Residual Networks (ResNets) \cite{He2015,He2016}, Recurrent Neural Networks (RNNs) and Long Short-Term Memory (LSTM) networks  \cite{Hochreiter1997}, and Auto-encoders \cite{Bengio2009} that are used frequently in practice. It turns out that our analysis -- which is based on the network's JM -- can also be easily extended to such DNN architectures. In fact, the proposed framework encompasses all DNN architectures for which the JM can be computed.
Below we compute the JM of a ResNet.

The ResNets introduce shortcut connection between layers. In particular, let $\phi(\cdot, \theta_l)$ denote a concatenation of several non-linear layers (see \eqref{eq:DNNlayer}). The $l$-th block of a Residual Network is then given as
\begin{IEEEeqnarray}{rCl}
 	\bz^l = \bz^{l-1} + \phi(\bz^{l-1}, \theta_{l}) \,.
\end{IEEEeqnarray}
We denote by $\bJ_l(\bz^{l-1})$ the JM of $\phi(\bz^{l-1}, \theta_{l})$. Then the JM of the $l$-th block is
\begin{IEEEeqnarray}{rCl}
	\frac{d \bz^l}{d \bz^{l-1}} = \bI + \bJ_l(\bz^{l-1})  \,,
\end{IEEEeqnarray}
and the JM of a ResNet is of the form
\begin{IEEEeqnarray}{rCl}
	\bJ_{SM}(\bz^{L-1}) \cdot \left(\bI + \sum_{l=1}^L \bJ_l(\bz^{l-1}) \left( \prod_{i=1}^{l-1} (\bI + \bJ_{l-i}(\bz^{l-2})) \right)  \right) \,, \nonumber \\ \* \label{eq:ResNetJM}
\end{IEEEeqnarray}
where $\bJ_{SM}(\bz^{L-1})$ denotes the JM of the soft-max layer. In particular, the right element of the product in \eqref{eq:ResNetJM} can be expanded as
\begin{IEEEeqnarray}{rll}
 && \bI \nonumber + \bJ_1(\bx) \nonumber \\
& & + \bJ_2(\bz^{1}) +  \bJ_2(\bz^{1})  \bJ_1(\bx)  \nonumber  \\
& & + \bJ_3(\bz^{2}) +  \bJ_3(\bz^{2}) \bJ_2(\bz^{1})  \bJ_1(\bx) + \bJ_3(\bz^{2})  \bJ_2(\bx) + \bJ_3(\bx)  \bJ_1(\bx)  \nonumber \\
& & + \ldots \nonumber  
\end{IEEEeqnarray}
This is a sum of JMs of all the possible sub-networks of a ResNet. In particular, there are $L$ elements of the sum consiting only of one 1-layer sub-networks and there is only one element of the sum consisting of $L$-layer sub-network. This observation is consistent with the claims in \cite{Veit2016b}, which states that ResNets resemble an ensemble of relatively shallow networks.

\subsection{Beyond the Euclidean Metric}

Moreover, we can also consider the geodesic distance on a manifold as a measure for margin instead of the Euclidean distance. The geodesic distance can be more appropriate than the Euclidean distance since it is a natural metric on the manifold. Moreover, the covering number of the manifold $\sX$ may be smaller if we use the covering based on the geodesic metric balls, which will lead to tighter GE bounds. We outline the approach below.

Assume that $\sX$ is a Riemannian manifold and $\bx, \bx' \in \sX$. Take a continuous, piecewise continuously differentiable curve $c(t)$, $t = [0,1]$ such that $c(0) = \bx$, $c(1) = \bx'$ and $c(t) \in \sX$ $\forall t \in [0,1]$. The set of all such curves $c(\cdot)$ is denoted by $\sC$. Then the geodesic distance between $\bx$ and $\bx'$ is defined as
\begin{IEEEeqnarray}{rCl}
	d_G(\bx, \bx') = \inf_{c(t) \in \sC} \int_0^1 \left \| \frac{d c(t)}{d t}  \right\|_2 d t	 \,.
\end{IEEEeqnarray}
Similarly as in Section~\ref{sec:DNNproperties}, we can show that the JM of DNN is central to bounding the distance expansion between the signals at the DNN input and the signals at the DNN output.
\begin{theorem} \label{th:distance_expansion_manifold}
	Take $\bx, \bx' \in \sX$, where $\sX$ is the Riemmanian manifold and take $c(t)$, $t = [0,1]$ to be a continuous, piecewise continuously differentiable curve connecting $\bx$ and $\bx'$ such that $d_G(\bx, \bx') =  \int_0^1 \left \| \frac{d c(t)}{d t}  \right\|_2 d t	$. Then
\begin{IEEEeqnarray}{rCl}
 \| f(\bx') - f(\bx) \|_2 \leq \sup_{t \in [0,1]} \| \bJ(c(t)) \|_2 d_G(\bx', \bx) 
\end{IEEEeqnarray}
\begin{IEEEproof}
The proof appears in Appendix~\ref{sec:th:proof:distance_expansion_manifold}.
\end{IEEEproof}
\end{theorem}
Note that we have established a relationship between the Euclidean distance of two points in the output space and the corresponding geodesic distance in the input space. This is important because it implies that promoting a large Euclidean distance between points can lead to a large geodesic distance between the points in the input space. Moreover, the ratio between $\|f(\bx') - f(\bx)\|_2$ and $d_G(\bx, \bx')$ is upper bounded by the maximum value of the spectral norm of the network's JM evaluated on the line $c(t)$. This result is analogous to the results of Theorem~\ref{th:jacobian_difference} and Corollary~\ref{th:dist_inequality}. It also implies that regularizing the network's JM as proposed in Section~\ref{sec:margin_bounds} is beneficial also in the case when the classification margin is not measured in the Euclidean metric.

Finally, note that in practice the training data may not be balanced. The provided GE bounds are still valid in such cases. However, the classification error may not the best measure of performance in such cases as it is dominated by the classification error of the class with the highest prior probability. Therefore, alternative performance measures need to be considered. We leave a detailed study of training DNN with unbalanced training sets for possible future work.

\section{Experiments} \label{sec:experiments}

We now validate the theory with a series of experiments on the MNIST \cite{LeCun1998}, CIFAR-10 \cite{Krizhevsky2009}, LaRED \cite{Hsiao2014} and ImageNet (ILSVRC2012) \cite{ILSVRC15} datasets. The Jacobian regularizer is applied to various DNN architectures such as DNN with fully connected layers, convolutional DNN and ResNet \cite{He2015}. We use the ReLUs in all considered DNNs as this is currently the most popular non-linearity.


\subsection{Fully Connected DNN}

In this section we compare the performance of fully connected DNNs regularized with Jacobian Regularization or with the weight decay. Then we analyse the behaviour of the JM of a fully connected DNNs of various depth and width.

\subsubsection{Comparison of Jacobian Regularization and Weight Decay}

\begin{figure}[t]
	\centering
	\subfigure[MNIST]{
	\includegraphics[height=0.15\paperwidth]{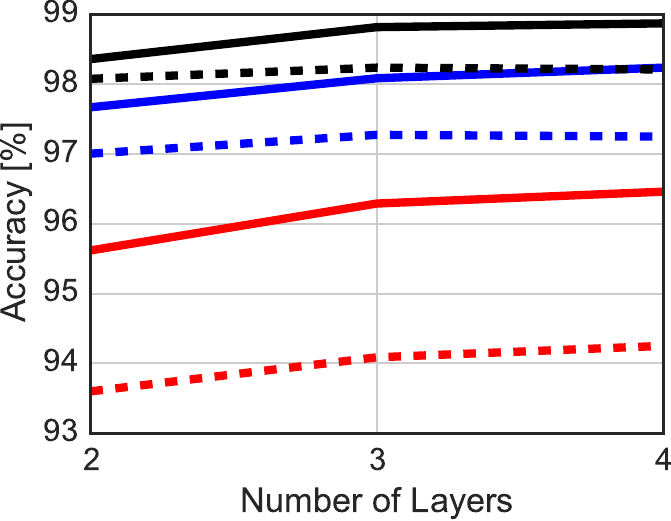}}	
	\subfigure[CIFAR-10]{
	\includegraphics[height=0.15\paperwidth]{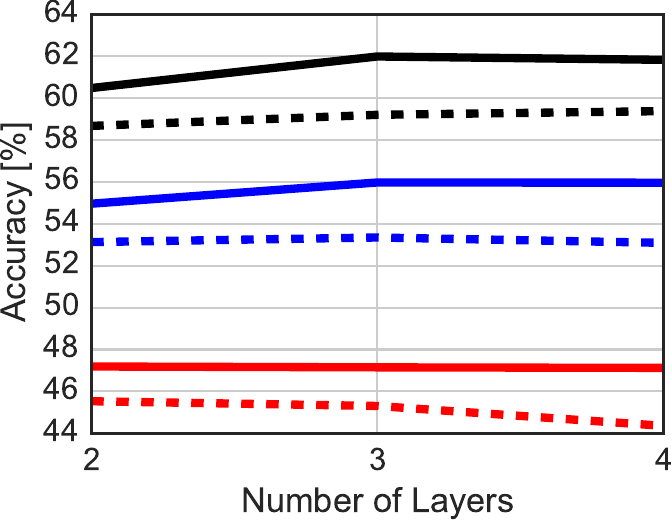}}
\caption{Classification accuracy of DNNs trained with the Jacobian regularization (solid lines) and the weight decay (dashed lines). Different numbers of training samples are used: 5000 (red), 20000 (blue) and 50000 (black).} \label{fig:mnist_cifar_dnn}	
\vspace{-0.15cm}
\end{figure}

First, we compare standard DNN with fully connected layers trained with the weight decay and the Jacobian regularization \eqref{eq:regF} on the MNIST and CIFAR-10 datasets. Different number of training samples are used (5000, 20000, 50000). We consider DNNs with 2, 3 and 4 fully connected layers where all layers, except the last one, have dimension equal to the input signal dimension, which is 784 in case of MNIST and 3072 in case of CIFAR-10. The last layer is always the softmax layer and the objective is the CEE loss. The networks were trained using the stochastic gradient descent (SGD) with momentum, which was set to 0.9. Batch size was set to 128 and learning rate was set to 0.01 and reduced by factor 10 after every 40 epochs. The networks were trained for 120 epochs in total. The weight decay and the Jacobian regularization factors were chosen on a separate validation set. The experiments were repeated with the same regularization parameters on 5 random draws of training sets and weight matrix initializations. Classification accuracies averaged over different experimental runs are shown in Fig.~\ref{fig:mnist_cifar_dnn}.
We observe that the proposed Jacobian regularization always outperforms the weight decay. This validates our theoretical results in Section~\ref{sec:margin_bounds}, which predict that the Jacobian matrix is crucial for the control of (the bound to) the GE. Interestingly, in the case of MNIST, a 4 layer DNN trained with 20000 training samples  and Jacobian regularization (solid blue line if Fig. \ref{fig:mnist_cifar_dnn} (a)) performs on par with DNN trained with 50000 training samples and weight decay (dashed black line Fig. \ref{fig:mnist_cifar_dnn} (a)), which means that the Jacobian regularization can lead to the same performance with significantly less training samples.

\subsubsection{Analysis of Weight Normalized Deep Neural Networks} \label{sec:experiment_weight_normalization}

Next, we explore weight normalized DNNs, which are described in Section~\ref{sec:margin_bounds}. We use the MNIST dataset and train DNNs with a different number of fully connected layers ($L = 2,3,4, 5$) and different sizes of weight matrices ($M_l =  784,2\cdot 784, 3\cdot 784, 4\cdot 784, 5\cdot 784, 6\cdot 784$, \mbox{$l = 1,\ldots, L-1$)}. The last layer is always the softmax layer and the objective is the CCE loss.  The networks were trained using the stochastic gradient descent (SGD) with momentum, which was set to 0.9. Batch size was set to 128 and learning rate was set to 0.1 and reduced by factor 10 after every 40 epochs. The networks were trained for 120 epochs in total. All experiments are repeated 5 times with different random draws of a training set and different random weight initializations. We did not employ any additional regularization as our goal here is to explore the effects of the weight normalization on the DNN behaviour. We always use 5000 training samples.

\begin{figure*}[t]
	\centering
	\subfigure[Classification accuracy.]{\includegraphics[height=0.15\paperwidth]{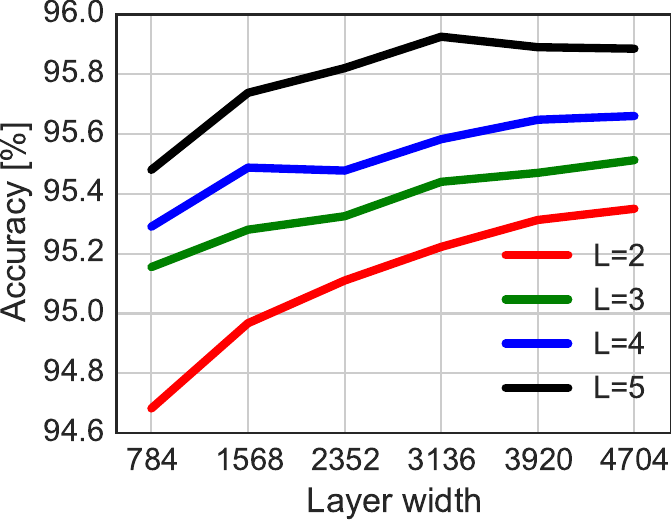}} 
	\hspace{0cm}\subfigure[Smallest $o(s_i)$ (training set).]{\includegraphics[height=0.15\paperwidth]{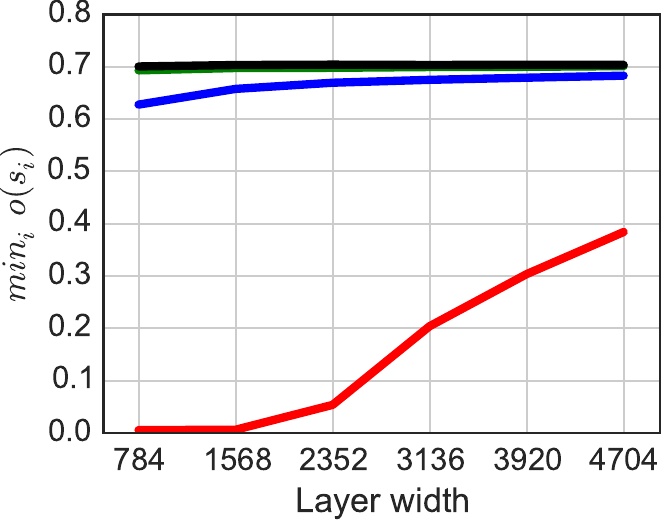}} 
	\subfigure[Largest $\| \bJ(\bx_i) \|_2$ (training set).]{\includegraphics[height=0.15\paperwidth]{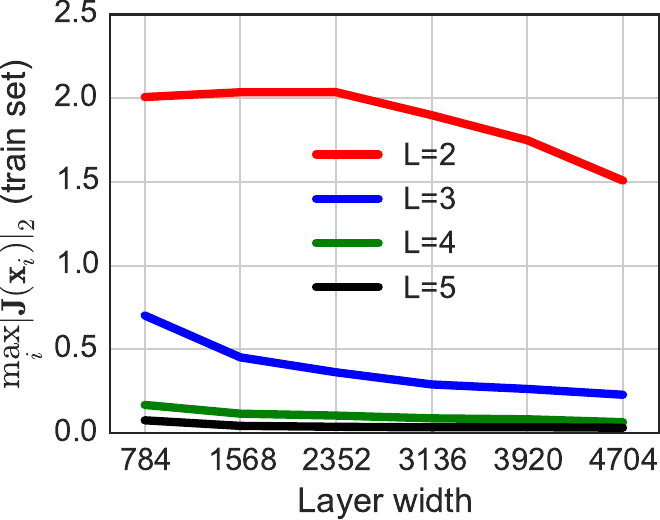}}
		\hspace{0cm}
	\subfigure[Largest $\| \bJ(\bx_i) \|_2$ (test set).]{\includegraphics[height=0.15\paperwidth]{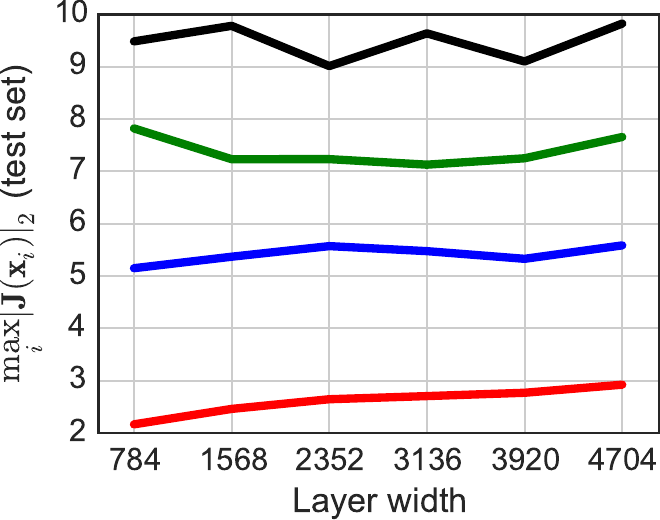}}	
\caption{Weight normalized DNN with $L=2,3,4,5$ layers and different sizes of weight matrices (layer width). Plot (a) shows classification accuracy, plot (b) shows the smallest score of training samples, plot (c) shows the largest spectral norm of the network's JM evaluated on the training set and plot (d) shows the largest spectral norm of the network's JM evaluated on the testing set.} \label{fig:weight_normalization}	
\vspace{-0.15cm}
\end{figure*}

The classification accuracies are shown in Fig. \ref{fig:weight_normalization} (a) and the smallest classification score obtained on the training set is shown in Fig. \ref{fig:weight_normalization} (b). We have observed for all configurations that the training accuracies were 100\% (only exception is the case $L=2$, $M_l = 784$ where the training accuracy was 99.6\%). Therefore, the (testing set) classification accuracies increasing with the network depth and the weight matrix size directly imply that the GE is smaller for deeper and wider DNNs. Note also that the score increases with the network depth and width. This is most obvious for the 2 and 3 layer DNNs, whereas for the 3 and 4 layer DNNs the score is close to $\sqrt{2}$ for all network widths. 

Since the DNNs are weight normalized, the Frobenious norms of the weight matrices are equal to the square root of the weight matrix dimension, and the product of Frobenious norms of the weight matrices grows with the network depth and the weight matrix size. The increase of score with the network depth and network width does not offset the product of Frobenious norms, and clearly, the bound in \eqref{eq:GEbound_manifold_DNN} based on the margin bound in \eqref{eq:margin_bound_euc4} and the bound in \eqref{eq:erc_bahaviour}, which leverage the Frobenious norms of the weight matrices, predict that the GE will increase with the network depth and weight matrix size in this scenario. Therefore, the experiment indicates that these bounds are too pessimistic.

We have also inspected the spectral norms of the weight matrices of the trained networks. In all cases the spectral norms were greater than one. We can argue that the bound in \eqref{eq:GEbound_manifold_DNN} based on the margin bound in \eqref{eq:margin_bound_euc3} predicts that the GE will increase with network depth, as the product of the spectral norms grows with the network depth in a similar way than in previous paragraph. We  note however, that the spectral norms of the weight matrices are much smaller than the Frobenious norms of the weight matrices. 

Finally, we look for a possible explanation for the success of the weight normalization in the bounds in \eqref{eq:GEbound_manifold_DNN} based on the margin bounds in \eqref{eq:margin_bound_euc1} and \eqref{eq:margin_bound_euc2}, which are a function on the JM. The largest value of the spectral norm of the network's JM evaluated on the training set is shown in \mbox{Fig.~\ref{fig:weight_normalization} (c)} and the largest value of the spectral norm of the network's JM evaluated on the testing set is shown in \mbox{Fig.~\ref{fig:weight_normalization} (d)}. 

We can observe an interesting phenomena. The maximum value of the JM's spectral norm on the training set decreases with the network depth and width. On the other hand, the  maximum value of the JM's spectral norm on the testing set increases with network depth (and slightly with network width). From the perspective of the constraint sets in \eqref{eq:cset1} and \eqref{eq:cset2} we note that in the case of the latter we have to take into account the worst case spectral norm of the JM for inputs in $\text{conv}(\sX)$. The maximum value of the spectral norm on the testing set indicates that this value increases with the network depth and implies that the bound based on \eqref{eq:margin_bound_euc2} is still loose. On the other hand, the bound in \eqref{eq:margin_bound_euc1} implies that we have to consider the JM in the neighbourhood of the training samples. As an approximation, we can take the spectral norms of the JMs evaluated at the training set. As it is shown in \mbox{Fig.~\ref{fig:weight_normalization} (c)} this values decrease with the network depth and width. We argue that this provides a reasonable explanation for the good generalization of deeper and wider weight normalized DNNs.

\subsection{Convolutional DNN}

In this section we compare the performance of convolutional DNNs regularized with the Jacobian regularizer or with the weight decay. We also show that Jacobian Regularization can be applied to batch normalized DNNs. We will use the standard MNIST and CIFAR-10 dataset and the LaRED dateset which is briefly described below.

\review{
The LaRED dataset contains depth images of 81 distinct hand gestures performed by 10 subjects with approximately 300 images of each gesture per subject. We extracted the depth images of the hands using the masks provided in\cite{Hsiao2014} and resized the images to $32 \times 32$. The images of the  first 6 subjects were used to create non-overlapping training and testing sets. In addition we also constructed a testing set composed from the images of the last 4 subjects in the dataset in order to test generalization across different subjects. The goal is classification of gestures based on the depth image.}

\subsubsection{Comparison of Jacobian Regularization and Weight Decay} \label{sec:CNNmnist_lared}

We use a 4 layer convolutional DNN with the following architecture: $(32,5,5)$-conv, $(2,2)$-max-pool, $(32,5,5)$-conv, $(2,2)$-max-pool followed by a softmax layer, where $(k,u,v)$-conv denotes the convolutional layer with $k$ filters of size $u \times v$ and $(p,p)$-max-pool denotes max-pooling with pooling regions of size $p \times p$. The training procedure follows the one described in the previous paragraphs. The results are reported in Table~\ref{tab:mnist_lared_results}.

We observe that training with the Jacobian regularization outperforms the weight decay in all cases. This is most obvious at smaller training set sizes. For example, on the MNIST dataset, the DNN trained using 1000 training samples and regularized with the weight decay achieves classification accuracy of 94\% and the DNN trained with the Jacobian regularization achieves classification accuracy of 96.3\%.

\begin{table*}[t]
\scriptsize
\center
\caption{Classification acc. $[\%]$ of convolutional DNN on MINST and LaRED.} \label{tab:mnist_lared_results}
\begin{tabular}{ccccccccc}
    \toprule 
  \multicolumn{3}{c}{MNIST}   & \multicolumn{3}{c}{LaRED (same subject)} &  \multicolumn{3}{c}{LaRED (different subject)}\\ \cmidrule(r){1-3}   \cmidrule(r){4-6}   \cmidrule(r){7-9}   
\# train samples & Weight dec. & Jac. reg. & \# train samples & Weight dec. & Jac. reg. & \# train samples & Weight dec. & Jac. reg.  \\
\midrule
1000 & 94.00  & \textbf{96.03} & 2000 & 61.40  & \textbf{63.56} & 2000 & 31.53  & \textbf{32.62}  \\
5000 &  97.59 & \textbf{98.20} & 5000 &  76.59 & \textbf{79.14} & 5000 &  38.11 & \textbf{39.62} \\
20000 & 98.60 & \textbf{99.00} & 10000 & 87.01 & \textbf{88.24} & 10000 & 41.18 & \textbf{42.85} \\
50000 & 99.10 & \textbf{99.35} & 50000 & 97.18 & \textbf{97.54} & 50000 & 45.12 & \textbf{46.78} \\
 \bottomrule
\end{tabular} 
\vspace{-0.0cm}
\end{table*}

\review{
Similarly, on the LaRED dataset the Jacobian regularization outperforms the weight decay with the difference most obvious at the smallest number of training samples. Note also that the generalization of the network to the subjects outside the training set is not very good; i.e., using 50000 training samples the classification accuracy on the testing set containing the same subjects is higher than 97\% whereas the classification accuracy on the testing set containing different subjects is only 46\%. Nevertheless, the Jacobian regularization outperforms the weight decay also on this testing set by a small margin.}

\subsubsection{Batch Normalization and Jacobian Regularization}

Now we show that the Jacobian regularization \eqref{eq:regF} can also be applied to a batch normalized DNN. Note that we have shown in Section~\ref{sec:margin_bounds} that the batch normalization has an effect of normalizing the rows of the weight matrices. 

We us the CIFAR-10 dataset and use the All-convolutional-DNN proposed in \cite{Springenberg2015} (All-CNN-C) with 9 convolutional layers, an average pooling layer and a softmax layer. All the convolutional layers are batch normalized and the softmax layer is weight normalized. The networks were trained using the stochastic gradient descent (SGD) with momentum, which was set to 0.9. Batch size was set to 64 and the learning rate was set to 0.1 and reduced by a factor 10 after every 25 epochs. The networks were trained for 75 epochs in total. The classification accuracy results are presented in Table~\ref{tab:allcnn_results} for different sizes of training sets (2500, 10000, 50000).

We can observe that the Jacobian regularization also leads to a smaller GE in this case. 

\begin{table}[t] 
\scriptsize
\center
\caption{Classification acc. $[\%]$ of convolutional DNN on CIFAR-10.} \label{tab:allcnn_results}
\begin{tabular}{ccc}
    \toprule
\# train samples & Batch norm. & Batch norm. + Jac. reg.  \\
\midrule
2500 & 60.86  & \textbf{66.15} \\
10000 &  76.35 & \textbf{80.57} \\
50000 & 87.44 & \textbf{88.95} \\
 \bottomrule
\end{tabular} 
\vspace{-0.0cm}
\end{table}

\subsection{Residual Networks}

Now we demonstrate that the Jacobian regularizer is also effective when applied to ResNets. We use the CIFAR-10 and ImageNet datasets. We use the per-layer Jacobian regularization \eqref{eq:regFefficient} for experiments in this section. 

\subsubsection{CIFAR-10}

The Wide ResNet architecture proposed in \cite{Zagoruyko2016}, which follows \cite{He2016}, but proposes wider and shallower networks which leads to the same or better performance than deeper and thinner networks is used here. In particular, we use the ResNet with 22 layers of width 5.

We follow the data normalization process of \cite{Zagoruyko2016}. We also follow the training procedure of \cite{Zagoruyko2016} except for the learning rate and use the learning rate sequence: (0.01, 5), (0.05, 20), (0.005, 40), (0.0005, 40), (0.00005, 20), where the first number in parenthesis corresponds to the learning rate and the second number corresponds to the number of epochs. We train ResNet on small training sets (2500 and 10000 training samples) without augmentation and on the full training set with the data augmentation as in \cite{Zagoruyko2016}. The regularization factor were set to 1 and $0.1$ for the smaller training sets (2500 and 10000) and the full augmented training set, respectively.

The results are presented in Table~\ref{tab:resnet_results}.  In all cases the ResNet with Jacobian regularization outperforms the standard ResNet. The effect of regularization is the strongest with the smaller number of training samples, as expected.

\begin{table}[t] 
\scriptsize
\center
\caption{Classification acc. $[\%]$ of ResNet CIFAR-10} \label{tab:resnet_results}
\begin{tabular}{ccc}
    \toprule
\# train samples & ResNet & ResNet + Jac. reg.  \\
\midrule
2500 & 55.69  & \textbf{62.79 } \\
10000  &  71.79 & \textbf{78.70} \\
50000 + aug. & 93.34 & \textbf{94.32} \\
 \bottomrule
\end{tabular} 
\vspace{-0.0cm}
\end{table}

\subsubsection{ImageNet} \label{sec:resnet_imagenet}
We use the 18 layer ResNet \cite{He2015} with identity connection \cite{He2016}. The training procedure follows \cite{He2015} with the learning rate sequence: (0.1, 30), (0.01, 30), (0.001, 30). The Jacobian regularization factor is set to 1.

\begin{figure*}[t]
	\centering
	\subfigure[No data augmentation.]{\includegraphics[height=0.15\paperwidth]{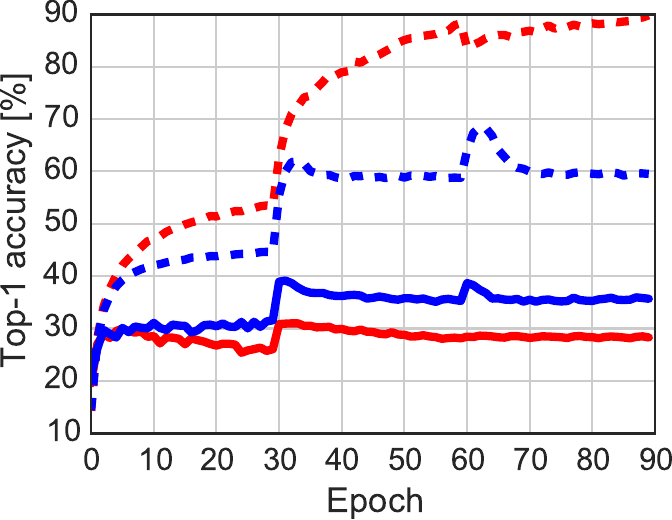}} 
	\subfigure[No data augmentation.]{\includegraphics[height=0.15\paperwidth]{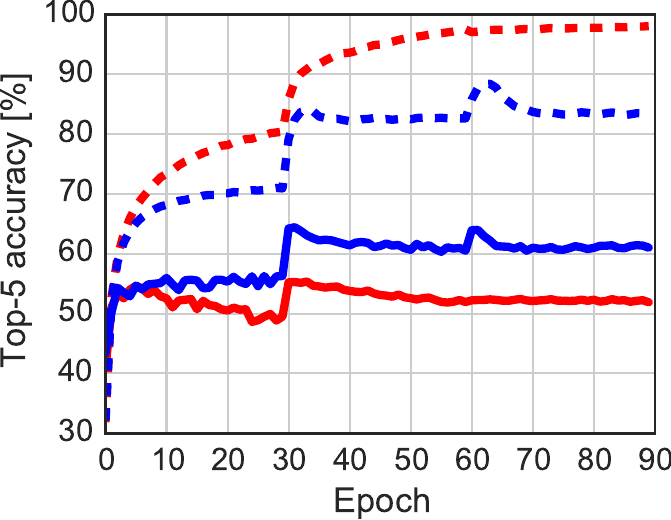}} 
	\subfigure[Data augmentation.]{\includegraphics[height=0.15\paperwidth]{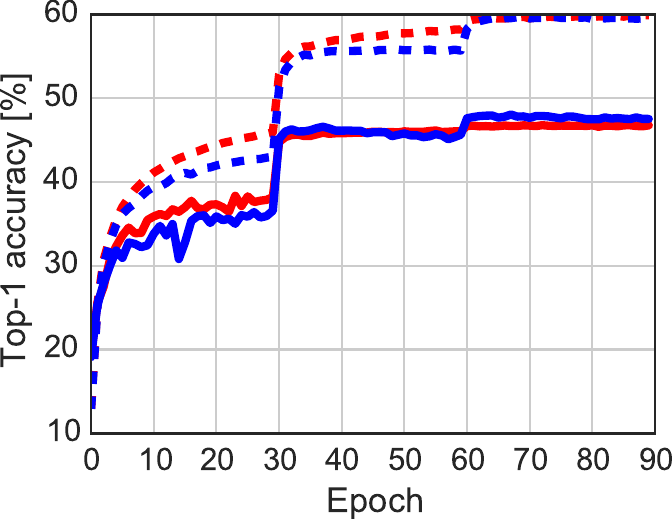}}
		\hspace{0cm}
	\subfigure[Data augmentation.]{\includegraphics[height=0.15\paperwidth]{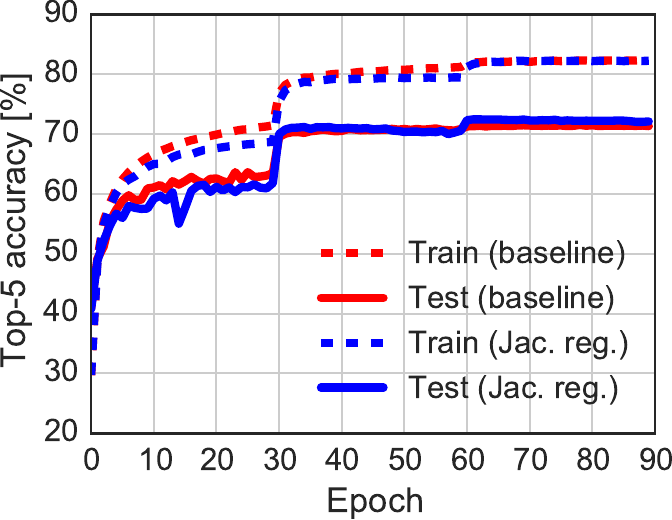}}	
\caption{\review{Training set (dashed) and testing set (solid) classification accuracies during training. Blue curves correspond to the ResNet with Jacobian regularization and red curves correspond to the baseline ResNet. Top-1 and top-5 classification accuracies are reported for training without data augmentation (a,b) and for training with data augmentation (c,d).}}
 \label{fig:imagenet_training}	
\vspace{-0.0cm}
\end{figure*}

The images in the dataset are resized to $128 \times 128$. We run an experiment without data augmentation and with data augmentation following \cite{Krizhevsky2012}, which includes random cropping of images of size $112 \times 112$ from the original image and color augmentation. The classification accuracies during training are shown in Fig.~\ref{fig:imagenet_training} and the final results are reported in Table~\ref{tab:imagenet_results}.

We first focus on training without data augmentation. The ResNet trained using the Jacobian regularization has a much smaller GE (23.83\%) compared to the baseline ResNet (61.53\%). This again demonstrates that the Jacobian regularization decreases the GE, as our theory predicts. Note that the smaller GE of Jacobian regularized ResNet partially transfers to a higher classification accuracy on the testing set. However, in practice DNNs are often trained with data augmentation. In this case the GE of a baseline ResNet is much lower (13.14\%) and is very close to the GE of the ResNet with the Jacobian regularization (12.03\%). It is clear that data augmentation reduces the need for strong regularization. Nevertheless, note that the ResNet trained with the Jacobian regularization achieves a slightly higher testing set accuracy (47.51\%) compared to the baseline ResNet (46.75\%).
\begin{table}[t] 
\scriptsize
\center
\caption{Classification acc. $[\%]$ and $\text{GE}$ $[\%]$ of ResNet on Imagenet.} \label{tab:imagenet_results}
\begin{tabular}{cccc}
    \toprule
 Setup & Train & Test  & $\text{GE}$  \\
\midrule
 Baseline  & 89.82 & 28.29 &  61.53 \\
 Baseline + Jac. reg. & 59.52 & 35.69 & 23.83 \\ 
 Baseline + aug. & 59.89 & 46.75 & 13.14 \\
 Baseline + aug + Jac. reg. & 59.54 & \textbf{47.51} & \textbf{12.03} \\ 
 \bottomrule
\end{tabular} 
\vspace{-0.0cm}
\end{table}

\subsection{Computational Time}

Finally, we measure how the use of Jacobian regularization affects training time of DNNs. We have implemented DNNs in Theano \cite{theano-short2016}, which includes automatic differentiation and computation graph optimization. The experiments are run on the Titan X GPU. The average computational time per batch for the convolutional DNN on the MNIST dataset in Section~\ref{sec:CNNmnist_lared} and for the ResNet on the ImageNet dataset in Section~\ref{sec:resnet_imagenet} are reported in Table~\ref{tab:timing_results}. Note that in the case of MNIST the regularizer in \eqref{eq:regF} is used and in the case of ImageNet the per-layer regularizer in \eqref{eq:regFefficient} is used. These results are also representative of the other datasets and network architectures.
\begin{table}[t] 
\scriptsize
\center
\caption{Average computation time  [$s$/batch]} \label{tab:timing_results}
\begin{tabular}{cccc}
    \toprule
Experiment & no reg. & Jac. reg. & Increase factor \\
\midrule
MNIST (Sec.\ref{sec:CNNmnist_lared}) & 0.003  & 0.030   & 10.00 \\
ImageNet (Sec. \ref{sec:resnet_imagenet}) & 0.120 & 0.190 & 1.580 \\
 \bottomrule
\end{tabular} 
\vspace{-0.0cm}
\end{table}

We can observe that using the Jacobian regularizer in \eqref{eq:regF} introduces additional computational time. This may not be critical if the number of training samples is small and training computational time is not too critical. On the other hand, the per-layer Jacobian regularizer in \eqref{eq:regFefficient} has a much smaller cost. As shown in the experiments this regularizer is still effective and leads to only 58\% increase in computation time on the ImageNet dataset. Due to its efficiency the per-layer Jacobian regularizer  might be more appropriate for large scale experiments where computational time is important.

\section{Conclusion} \label{sec:conclusions}

This paper studies the GE of DNNs based on their classification margin. In particular, our bounds express the generalization error as a function of the classification margin, which is bounded in terms of the achieved separation between the training samples at the network output and the network's JM.

One of the hallmarks of our bounds relates to the fact that the characterization of the behaviour of the generalization error is tighter than that associated with other bounds in the literature. Our bounds predict that the generalization error of deep neural networks can be independent of their depth and size whereas other bounds say that the generalization error is exponential in the network width or size.

Our bounds also suggest new regularization strategies such as the regularization of the network's Jacobian matrix, which can be applied on top of other modern DNN training strategies such as the weight normalization and the batch normalization, where the standard weight decay can not be applied.  These regularization strategies are especially effective in the limited training data regime in comparison to other approaches, with moderate increase in computational complexity.


\appendix

\subsection{Proof of Theorem \ref{th:jacobian_difference}} \label{sec:th:proof:jacobian_difference}

We first note that the line between $\bx$ and $\bx'$ is given by $\bx + t(\bx' - \bx)$, $t \in [0,1]$ . We define the function $F(t) = f( \bx + t (\bx' - \bx))$, and observe that $\frac{dF(t)}{dt} = \bJ(\bx + t (\bx' - \bx)) (\bx' -  \bx)$. By the generalized fundamental theorem of calculus or the Lebesgue differentiation theorem we write
\begin{IEEEeqnarray}{rCl}
	f(\bx')  - f(\bx) &=& F(1)  - F(0) = \int_{0}^1 \frac{dF(t)}{dt} \,dt \nonumber \\ &=& \int_{0}^1 \bJ(\bx + t (\bx' - \bx))  \,dt \, (\bx' -  \bx)\,.  \label{eq:jacobian_integral_proof}
\end{IEEEeqnarray}
This concludes the proof.

\subsection{Proof of Corollary \ref{th:dist_inequality}} \label{sec:th:proof:dist_inequality}
First note that $\|\bJ_{\bx, \bx'} (\bx' - \bx) \|_2 \leq \|\bJ_{\bx, \bx'} \|_2  \|\bx' - \bx \|_2$ and that $\bJ_{\bx, \bx'}$ is an integral of $\bJ(\bx + t (\bx' - \bx))$. In addition, notice that we may always apply the following upper bound:
\begin{IEEEeqnarray}{rCl}
	\| \bJ_{\bx, \bx'} \|_2 \leq \sup_{\bx, \bx' \in \sX, t \in [0,1]} \|\bJ(\bx + t (\bx' - \bx))\|_2 \,.
\end{IEEEeqnarray}
Since $\bx + t (\bx'-\bx) \in \text{conv}(\sX)$ $\forall t \in [0,1]$, we get \eqref{eq:euc_dist_inequality}.

\subsection{Proof of Lemma \ref{th:jacobian_spectral_norm_bound}} \label{sec:th:proof:jacobian_spectral_norm_bound}

In all proofs we leverage the fact that for any two matrices $\bA$, $\bB$ of appropriate dimensions it holds $\| \bA \bB \|_2 \leq \| \bA \|_2 \| \bB \|_2$. We also leverage the bound $\| \bA \|_2 \leq \| \bA \|_F$.

We start with the proof of statement 1). For the non-linear layer \eqref{eq:DNNlayer}, we note that the JM is a product of a diagonal matrix \eqref{eq:JM_nonlinear_diag} and the weight matrix $\bW_l$. Note that for all the considered non-linearities the diagonal elements of \eqref{eq:JM_nonlinear_diag} are bounded by 1 (see derivatives in Table~\ref{tab:nonlinearities}), which implies that the spectral norm of this matrix is bounded by 1. Therefore the spectral norm of the JM is upper bounded by $\| \bW_l\|_2$. The proof for the linear layer is trivial. In the case of the softmax layer \eqref{eq:softmax_layer} we have to show that the spectral norm of the softmax function $\left( - \zeta(\hat{\bz}) \zeta(\hat{\bz})^T + \diag(\zeta(\hat{\bz}) \right)$ is bounded by 1. We use the Gershgorin disc theorem, which states that the eigenvalues of $\left( - \zeta(\hat{\bz}) \zeta(\hat{\bz})^T + \diag(\zeta(\hat{\bz}) \right)$ are bounded by 
\begin{IEEEeqnarray}{rCl}	
	\max_i (\zeta(\hat{\bz}))_i (1- (\zeta(\hat{\bz}))_i) + (\zeta(\hat{\bz}))_i \sum_{j\neq i} (\zeta(\hat{\bz}))_j \,.
\end{IEEEeqnarray}
Noticing that $\sum_{j\neq i} (\zeta(\hat{\bz}))_j \leq 1$ leads to the upper bound 
\begin{IEEEeqnarray}{rCl}	
\textstyle
	\max_i (\zeta(\hat{\bz}))_i (2- (\zeta(\hat{\bz}))_i) \,. \label{eq:proof:temp1}
\end{IEEEeqnarray}
Since $(\zeta(\hat{\bz}))_i \in [0,1]$ it is trivial to show that \eqref{eq:proof:temp1} is upper bounded by 1.

The proof of statement 2) is straightforward. Because the pooling regions are non-overlapping it is straightforward to verify that the rows of all the defined pooling operators $\bP^l(\bz^{l-1})$ are orthonormal. Therefore, the spectral norm of the JM is equal to 1.

\subsection{Proof of Theorem \ref{th:margin_bound_euc}} \label{sec:th:proof:margin_bound_euc}

Throughout the proof we will use the notation $o(s_i) = o(\bx_i, y_i)$ and $\bv_{i j} = \sqrt{2}(\boldsymbol{\delta}_{i} - \boldsymbol{\delta}_j)$. We start by proving the inequality in \eqref{eq:margin_bound_euc1}. Assume that the classification margin $\gamma^d(s_i)$ of training sample $(\bx_i, y_i)$ is given and take $j^\star = \argmin_{j \neq y_i} \min \bv_{y_i j}^T f(\bx_i)$. We then take a point $\bx^\star$ that lies on the decision boundary between $y_i$ and $j^\star$ such that $o(\bx^\star, y_i) = 0$. Then 
\begin{IEEEeqnarray}{rCl}
	o(\bx_i,y_i) &=& o(\bx,y_i) - o(\bx^\star,y_i) = \bv^T_{y_i j^\star} (f(\bx_i) - f(\bx^\star)) \nonumber \\
	&=& \bv^T_{y_i j^\star} \bJ_{\bx_i,\bx^\star} (\bx_i - \bx^\star) \leq \| \bJ_{\bx_i,\bx^\star} \|_2 \|\bx_i - \bx^\star \|_2. \nonumber 
\end{IEEEeqnarray}
Note that by the choice of $\bx^\star$, $\|\bx_i - \bx^\star \|_2 = \gamma^d(s_i)$ and similarly $\| \bJ_{\bx_i,\bx^\star} \|_2 \leq \sup_{\bx: \|\bx - \bx_i\|_2 \leq \gamma^{d}(s_i)} \left \|  \bJ(\bx) \right \|_2$. Therefore, we can write
\begin{IEEEeqnarray}{rCl}
o(s_i) \leq \sup_{\bx: \|\bx - \bx_i\|_2 \leq \gamma^{d}(s_i)} \left \|  \bJ(\bx) \right \|_2 \, \gamma^d(s_i) ,
\end{IEEEeqnarray}
which leads to \eqref{eq:margin_bound_euc1}.

Next, we prove \eqref{eq:margin_bound_euc2}. Recall the definition of the classification margin in \eqref{eq:input_margin}:
\begin{IEEEeqnarray}{rCl}
 \gamma^{d}(s_i) &=& \sup \{ a : \|\bx_i - \bx\|_2 \leq a  \implies g(\bx) = y_i \, \forall \bx\} \nonumber \\ &=& \sup \{ a :  \|\bx_i - \bx \|_2 \leq a  \implies o(\bx,y_i) > 0 \, \forall \bx\} \,, \nonumber 
\end{IEEEeqnarray}
where we leverage the definition in \eqref{eq:output_score}. We observe 
$o(\bx,y_i) > 0 \iff  \min_{j\neq y_i} \bv^T_{y_i j} f(\bx) > 0$ and 
\begin{IEEEeqnarray}{C}
	\min_{j\neq y_i} \bv^T_{y_i j} f(\bx) = \min_{j\neq y_i} \left( \bv_{y_i j}^T f(\bx_i)  +  \bv^T_{y_i j}(f(\bx)-f(\bx_i))\right) \,. \nonumber 
\end{IEEEeqnarray}
Note that
\begin{IEEEeqnarray}{rCl}
&&\min_{j\neq y_i} \left( \bv^T_{y_i j} f(\bx_i) +  \bv^T_{y_i j}(f(\bx)-f(\bx_i)) \right)  \\ 
&& \geq \min_{j\neq y_i} \bv^T_{y_i j} f(\bx_i) + \min_{j\neq y_i}  \bv^T_{y_i j}(f(\bx)-f(\bx_i))  \nonumber \\ 
&& = o(\bx_i,y_i) + \min_{j\neq y_i}  \bv^T_{y_i j}(f(\bx)-f(\bx_i))\,.
\end{IEEEeqnarray}
Therefore, 
\begin{IEEEeqnarray}{rCl}
o(\bx_i, y_i)  +  \min_{j\neq y_i}  \bv_{y_i j}^T(f(\bx)-f(\bx_i)) > 0 \implies  o(\bx,y_i) > 0\,. \nonumber 
\end{IEEEeqnarray}
This leads to the bound of the classification margin
\begin{IEEEeqnarray}{rCl}
	\gamma^{d}(s_i) \geq \sup \{&& a : \| \bx_i - \bx \|_2 \leq a \implies  \nonumber \\
	&&  o(\bx_i,y_i)  +  \min_{j\neq y_i}  \bv_{y_i j}^T(f(\bx)-f(\bx_i)) > 0 \, \forall \bx\}  \,. \nonumber
\end{IEEEeqnarray}
Note now that 
\begin{IEEEeqnarray}{c}
	o(\bx_i, y_i)  +  \min_{j\neq y_i}  \bv_{y_i j}^T(f(\bx)-f(\bx_i)) > 0 \\
	 \iff  \nonumber \\
	 o(\bx_i, y_i)  -  \max_{j\neq y_i}  \bv_{y_i j}^T(f(\bx_i)- f(\bx)) > 0 \\
	 	 \iff  \nonumber \\
	o(\bx_i, y_i)  >  \max_{j\neq y_i}  \bv_{y_i j}^T(f(\bx_i)-f(\bx)) \,.
\end{IEEEeqnarray}
Moreover, 
\begin{IEEEeqnarray}{rCl}
\max_{j\neq y_i}  \bv_{y_i j}^T(f(\bx_i)-f(\bx)) \leq \sup_{\bx \in \text{conv}(\sX)} \| \bJ(\bx) \|_2 \| \bx_i - \bx \|_2 \,, \nonumber 
\end{IEEEeqnarray}
where we have leveraged the fact that $\| \bv_{ij} \|_2 = 1$ and the inequality \eqref{eq:euc_dist_inequality} in Corollary \ref{th:dist_inequality}. We may write
\begin{IEEEeqnarray}{rCl}
	\gamma^{d}(s_i) \geq \sup \{ a : && \| \bx_i - \bx \|_2 \leq a \implies  \nonumber \\
	 && o(\bx_i, y_i) > \sup_{\bx \in \text{conv}(\sX)} \| \bJ(\bx) \|_2  \| \bx_i - \bx \|_2 \, \forall \bx\}  . \nonumber 
\end{IEEEeqnarray}
$a$ that attains the supremum can be obtain easily and we get:
\begin{IEEEeqnarray}{rCl}
	\gamma^{d}(s_i) \geq  \frac{o(\bx_i, y_i)}{\sup_{\bx \in \text{conv}(\sX)} \| \bJ(\bx) \|_2}\,,
\end{IEEEeqnarray}
which proves \eqref{eq:margin_bound_euc2}. The bounds in \eqref{eq:margin_bound_euc3} and \eqref{eq:margin_bound_euc4} follow from the bounds provided in Lemma~\ref{th:jacobian_spectral_norm_bound} and the fact that the spectral norm of a matrix product is upper bounded by the product of the spectral norms. This concludes the proof.

\subsection{Proof of Theorem \ref{th:batchnorm}} \label{sec:batchnorm}
We denote by $\bW_l^N$ the row normalized matrix obtained from $\bW_l$ (in the same way as \eqref{eq:weight_normalization}). By noting that the ReLU and diagonal non-negative matrices commute, it is straight forward to verify that
\begin{IEEEeqnarray}{rCl}
\textstyle
	[\bN \left(\{\bz_i^l\}_{i=1}^m, \bW_l \right)\, \bW_l  \bz^{l}]_{\sigma} = \bN \left(\{\bz_i^l\}_{i=1}^m, \bW^N_l \right) [ \bW_l^N  \bz^{l}]_{\sigma}\,. \nonumber 
\end{IEEEeqnarray}
Note now that we can consider $\bN(\{\bz_i^l\}_{i=1}^m, \bW^N_l)$ as the part of the weight matrix $\bW_{l+1}$. Therefore, we can conclude that layer $l$ has row normalized weight matrix. When the batch normalization is applied to layers, all the weight matrices will be row normalized. The exception is the weight matrix of the last layer, which will be of the form $\bN(\{\bz_i^{L-1}\}_{i=1}^m, \bW_L) \bW_L$.

\subsection{Proof of Theorem \ref{th:distance_expansion_manifold}} \label{sec:th:proof:distance_expansion_manifold}

We begin by noting that $f(\bx') - f(\bx) = f(c(1)) - f(c(0))$ and
\begin{IEEEeqnarray}{c}
\textstyle
f(c(1)) - f(c(0)) =  \int_0^1 \frac{df(c(t))}{dt}\, dt = \int_0^1 \frac{df(c(t))}{dc(t)} \frac{dc(t)}{dt}\, dt \,, \nonumber
\end{IEEEeqnarray}
where the first equality follows from the generalized fundamental theorem of calculus, following the idea presented in the proof of Theorem~\ref{th:jacobian_difference}. The second equality follows from the chain rule of differentiation. Finally, we note that $\frac{df(c(t))}{c(t)} = \bJ(c(t))$ and that the norm of the integral is always smaller or equal to the integral of the norm and obtain
\begin{IEEEeqnarray}{rCl}
\textstyle
	\| f(\bx') - f(\bx) \|_2 &=& \left \| \int_0^1 \bJ(c(t)) \frac{dc(t)}{dt}\, dt \right \|_2 \nonumber \\
	&\leq &  \int_0^1 \|  \bJ(c(t)) \|_2  \left \| \frac{dc(t)}{dt} \right \|_2 \, dt \nonumber \\
	& \leq & \sup_{t \in [0,1]}  \|  \bJ(c(t)) \|_2  \int_0^1 \left \| \frac{dc(t)}{dt} \right \|_2 \, dt \nonumber \\
	&=& \sup_{t \in [0,1]}  \|  \bJ(c(t)) \|_2 d_G(\bx,\bx') \,,
\end{IEEEeqnarray}
where we have noted that $\int_0^1 \left \| \frac{dc(t)}{dt} \right \|_2 = d_G(\bx,\bx')$.  
\bibliographystyle{IEEEtran}
\bibliography{IEEEabrv,library_correct}
%

\end{document}